\documentclass[sigconf]{acmart}
\usepackage{amsmath}
\usepackage{url}
\usepackage{booktabs}
\usepackage{algorithm}
\usepackage{algpseudocode}
\usepackage{multirow}
\usepackage{graphicx}
\usepackage{subfigure}
\usepackage{cleveref}
\usepackage{subcaption}

\AtBeginDocument{%
  }

\setcopyright{acmlicensed}
\copyrightyear{2026}
\acmYear{2026}
\acmDOI{XXXXXXX.XXXXXXX}

\acmConference[Conference acronym 'XX]{Make sure to enter the correct
  conference title from your rights confirmation emai}{June 03--05,
  2018}{Woodstock, NY}
\acmISBN{978-1-4503-XXXX-X/18/06}

\begin{document}

\title{OrchMAS: Orchestrated Reasoning with Multi Collaborative\\Heterogeneous Scientific Expert Structured Agents}

\author{\texorpdfstring{%
  \textbf{Yichao Feng\textsuperscript{1,2}},
  \textbf{Haoran Luo\textsuperscript{2}\footnotemark[1]\authornote{Corresponding authors.}},
  \textbf{Zhenghong Lin\textsuperscript{2}},
  \textbf{Yiqun Sun\textsuperscript{1}},
  \textbf{Pengfei Wei\textsuperscript{1}}, \\
  \textbf{Lawrence B.~Hsieh\textsuperscript{1}},
  \textbf{Anh Tuan Luu\textsuperscript{2}\footnotemark[1]} \\
  \textsuperscript{1}Magellan Technology Research Institute (MTRI) \quad
  \textsuperscript{2}Nanyang Technological University, Singapore \\
  \texttt{\{yichao.feng, duke.sun, pengfei.wei, lawrence.hsieh\}@mtri.co.jp,}
  \texttt{haoran.luo@ieee.org,}
  \texttt{hongzhenglin970323@gmail.com,}
  \texttt{anhtuan.luu@ntu.edu.sg}
}{Yichao Feng, Zhenghong Lin, Yiqun Sun, Pengfei Wei, Lawrence B. Hsieh, Haoran Luo, Anh Tuan Luu}}

\AtBeginDocument{%
  \hypersetup{pdfauthor={Yichao Feng; Zhenghong Lin; Yiqun Sun; Pengfei Wei; Lawrence B. Hsieh; Haoran Luo; Anh Tuan Luu}}%
}

\makeatletter
\gdef\authors{Yichao Feng\and Haoran Luo\and Zhenghong Lin\and Yiqun Sun\and Pengfei Wei\and Lawrence B. Hsieh\and Anh Tuan Luu}
\gdef\shortauthors{Feng et al.}
\makeatother
\renewcommand{\shortauthors}{Feng et al.}

\begin{abstract}
Multi-agent large language model frameworks are promising for complex multi step reasoning, yet existing systems remain weak for scientific and knowledge intensive domains due to static prompts and agent roles, rigid workflows, and homogeneous model reliance, leading to poor domain adaptation, limited reasoning flexibility, and high latency on heterogeneous or long-horizon scientific tasks. They also struggle to revise earlier decisions when intermediate reasoning diverges, reducing reliability in structured and calculation heavy settings. To address these limitations, we propose a scientific domain oriented interactive two tier multi model orchestration framework. A dedicated orchestration model analyzes each task, dynamically constructs a domain aware reasoning pipeline, and instantiates specialized expert agents with tailored prompts, while an execution model performs each step under generated role and instruction specifications. The orchestrator iteratively updates the pipeline based on intermediate feedback, enabling dynamic replanning, role reallocation, and prompt refinement across multi turn interactions, strengthening robustness and specialization for scientific reasoning through structured heterogeneous model collaboration. The framework is model agnostic and supports heterogeneous LLM integration with different capacities or costs, enabling flexible performance efficiency trade offs in practical scientific deployments. Experiments show consistent improvements over existing multi agent systems and strong baselines across diverse reasoning and scientific style benchmarks. Our code is publicly available.\footnote{Github Code: \url{https://github.com/Githubuseryf/OrchMAS}}.
\end{abstract}



\maketitle
\begin{figure}[!t]
\centering
\includegraphics[width=\linewidth]{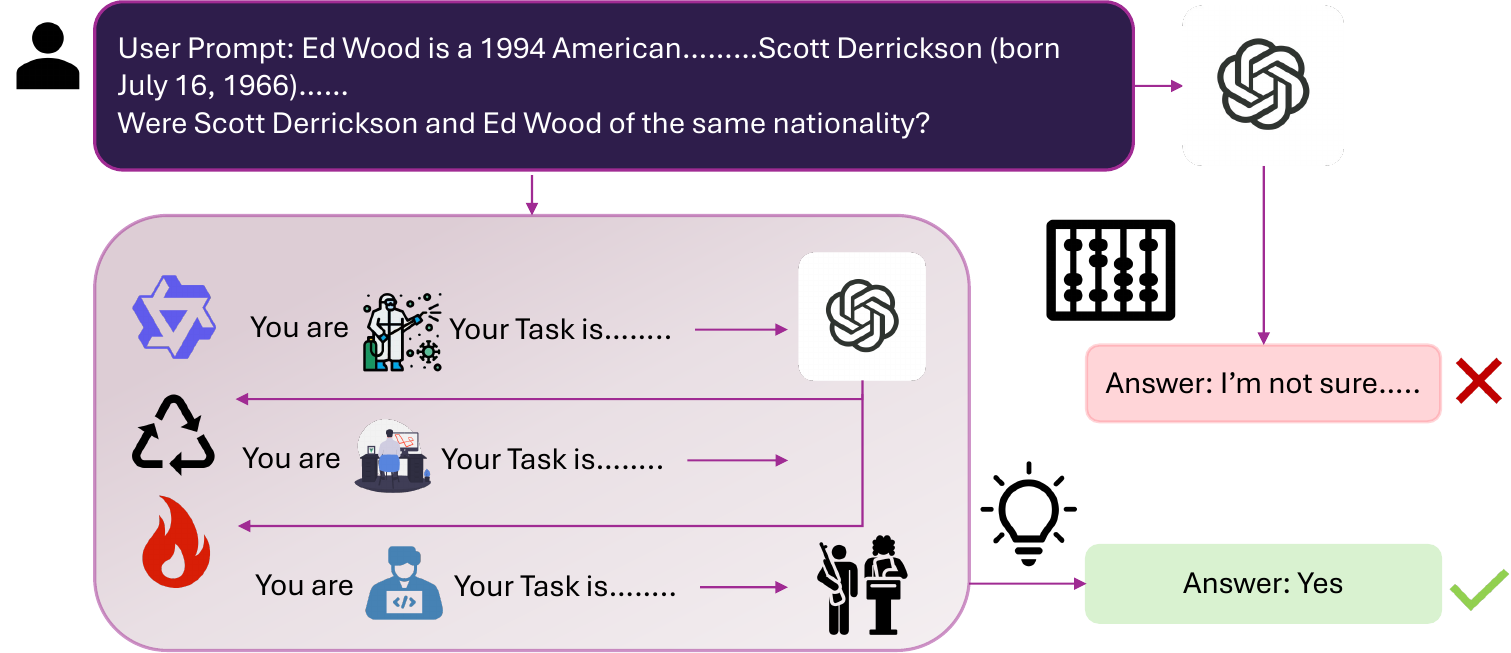}
\caption{Comparison of multi step and MAS LLM generation versus basic prompting.}
\label{fig:mas_page1}
\end{figure}
\section{Introduction}

\begin{figure*}[t]
\centering
\includegraphics[width=\linewidth]{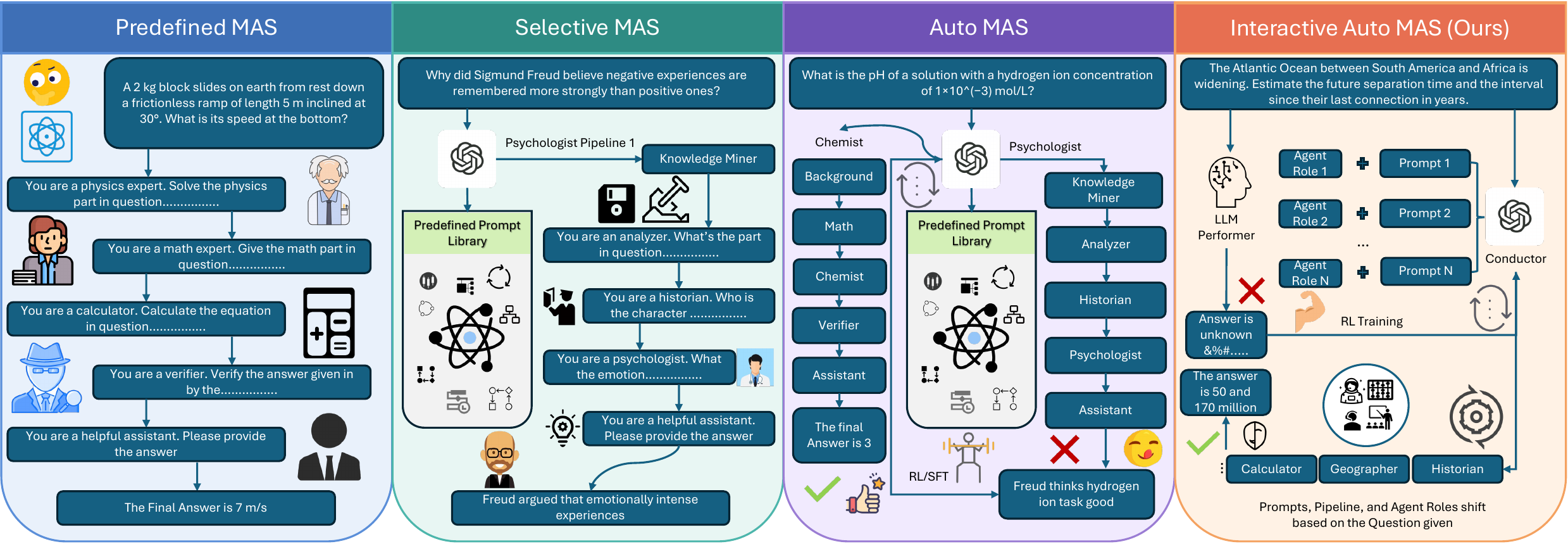}
\caption{Structural comparison of representative MAS frameworks. Predefined MAS and Selective MAS rely on static role templates and predefined agent libraries, while Auto MAS learns coordination through SFT/RL with a trained agent policy. Interactive Auto MAS performs multi turn dynamic agent and prompt generation and feedback guided orchestration.}
\label{fig:mas_page2}
\end{figure*}

Recent advances in large language models (LLMs) have demonstrated robust capabilities across many natural language processing (NLP) tasks~\citep{achiam2023gpt, luo2025hypergraphrag, wu2024affinity}. Nevertheless, when deployed as a lone reasoning entity, LLMs still exhibit fundamental limitations in accurately solving complex scientific tasks\citep{illusion-of-thinking}. First, they often struggle with logical and symbolic reasoning, failing to maintain consistency across extended chains of thought and algorithmic computations, especially in scientific subdomains that require numerical and symbolic calculation. Second, relying on a single model can create scalability, interpretability, and reliability bottlenecks when handling diverse subproblems or domain specific knowledge, since one monolithic model must represent all specialized reasoning strategies and planning behaviors~\citep{althaf2025multi, wu2024affinity}. Third, single LLM approaches remain prone to hallucinations, superficial pattern matching instead of causal inference, and sensitivity to irrelevant information, which undermines robustness and faithfulness in scientific~\citep{feng2025stimuli, feng2025aspect}. To address this, Multi-agent systems (MAS) have emerged as a promising paradigm for pipeline based reasoning and prompt level orchestration, as shown in Fig.~\ref{fig:mas_page1}, where specialized agents take distinct roles and collaboratively solve complex tasks through structured interaction~\citep{li2024survey}. MAS approaches able to coordinate multiple LLM based agents to decompose tasks into structured subtasks~\citep{talebirad2023multi}.

However, existing MAS still exhibit structural limitations that hinder their effectiveness on diverse scientific tasks~\citep{wang-etal-2025-megaagent, cemri2025multi}, as shown in Figure~\ref{fig:mas_page2}.
\textbf{Static prompts and fixed roles} remain common in prior MAS frameworks, where agents rely on predefined roles and generic templates within fixed pipelines~\citep{perera2025auto, xia2025parallelism, lu2024morphagent}. This \textbf{causes prompt/task misalignment} and weak domain adaptation, leading to brittle reasoning and missed domain specific strategies~\citep{tran2025multi, sun2025multi}. \textbf{High human cost workflows} are also widely used, with handcrafted multi stage pipelines and predetermined step orders. Such designs \textbf{incur high prompt engineering and maintenance cost}, reduce flexibility, and prevent dynamic stage reordering, verification insertion, or step skipping. They further \textbf{amplify error propagation}, since early mistakes are often inherited by downstream stages~\citep{su2025difficulty, chang2025sagallm}. \textbf{Single model role simulation} is another common pattern, where all agents are instantiated from the same LLM with different prompts. This \textbf{limits true specialization and verification reliability}: if the base model is biased or incomplete in a scientific domain, using it for both hypothesis generation and validation can compound errors. Such homogeneous setups \textbf{reduce reasoning diversity and cross agent complementarity}, weakening robustness on specialized tasks~\citep{yediversity, xu2026rethinking}.

In this paper, we propose \textbf{Orchestrated Multi-Agent System (OrchMAS)}, 
a task adaptive MAS reasoning framework for diverse scientific and 
knowledge intensive tasks. OrchMAS targets key limitations of existing MAS approaches, 
including static role design, high manual workflow cost, and homogeneous agent 
specialization. \textbf{First}, we introduce a \textbf{dynamic orchestration mechanism} for 
task aware pipeline construction and adaptive prompt generation, where 
task specific roles, prompts, and interaction structures are generated conditioned 
on the input problem to reduce prompt task misalignment and improve cross domain 
adaptability. \textbf{Second}, we design an \textbf{iterative and reconfigurable multi stage 
collaboration pipeline} with flexible stage ordering 
and intermediate result driven restructuring, enabling adaptive stage insertion and pruning to lower workflow engineering cost and limit error propagation. 
\textbf{Third}, we adopt a \textbf{two tier heterogeneous agent architecture} that separates 
high level planning from knowledge intensive inference, assigning different models 
to distinct responsibilities to improve specialization beyond 
homogeneous MAS designs.

The system dynamically customizes agent behavior according to problem domain and 
current answer confidence, and leverages execution feedback to trigger verification, 
plan revision, or stage skipping when necessary, improving reliability under uncertain intermediate states. 
We instantiate OrchMAS by training an action oriented orchestration agent with action based 
GRPO optimization~\citep{shao2024deepseekmath}, which supports stable policy learning, 
efficient credit assignment, and cross task transfer. Extensive experiments on multiple 
benchmarks demonstrate consistent improvements over strong baselines in both 
in-distribution and OOD settings, covering scientific QA, mathematical reasoning, 
and multi domain question answering across heterogeneous task formats. 

\begin{figure*}[t]
\centering
\includegraphics[width=\linewidth]{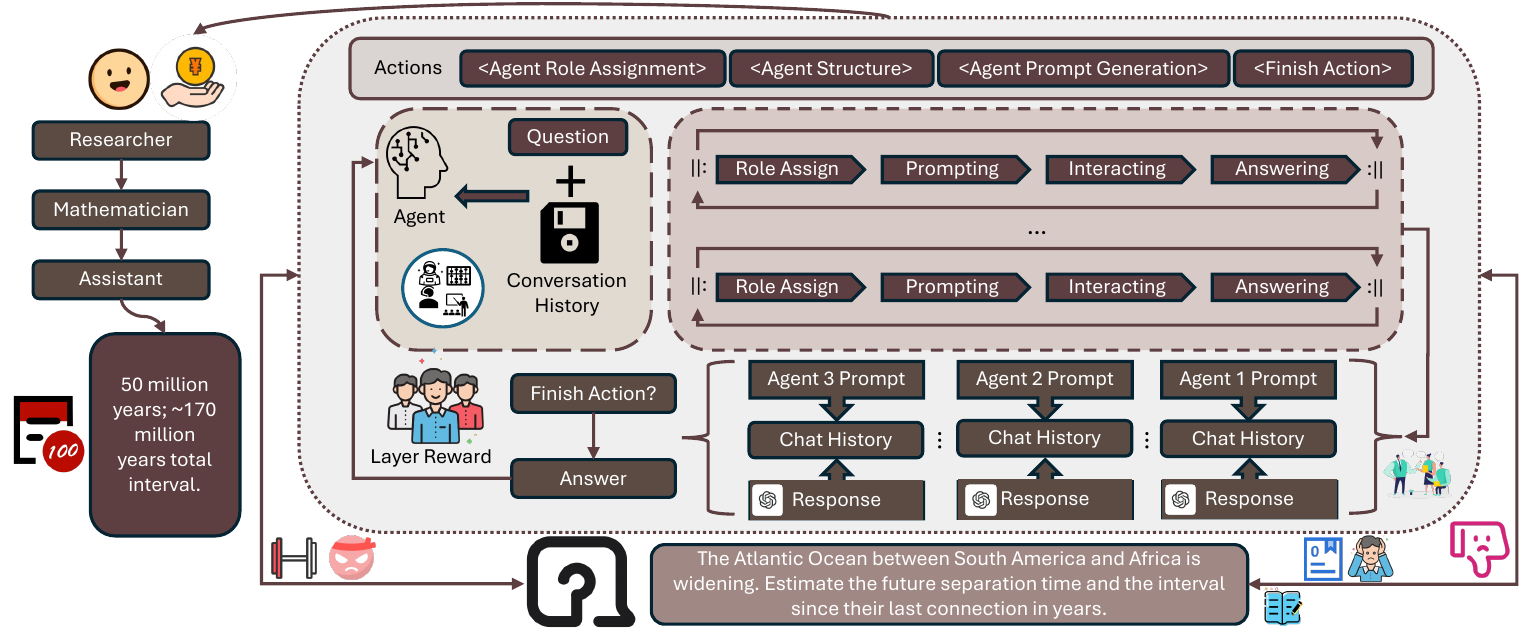}
\caption{Training workflow of the OrchMAS framework. The orchestration policy iteratively performs agent role assignment, agent structure construction, and agent prompt generation to dynamically instantiate task specific agents. The process repeats until a finish action is triggered. Layer wise rewards are computed from intermediate trajectories and final answers.}
\label{fig:mas_page3}
\end{figure*}

\section{Related Work}
In this section, we will introduce background and related works.

\textbf{LLM and MAS}
Recent years have witnessed rapid progress in LLM as general purpose reasoning engines~\citep{achiam2023gpt, peng2023instruction, AlphaGeo}. However, relying solely on simple prompting strategies often fails to consistently produce optimal or reliable answers, especially for complex multi step tasks~\citep{chan2023chateval}. To address this limitation, prior work has explored MAS paradigms built on structured, pipeline based collaboration. For example, AgentVerse~\citep{chen2023agentverse} introduces a pipeline oriented MAS framework and demonstrates strong empirical gains through coordinated agent interaction, works like MARS~\citep{zhang2025mars} used it for prompt enhancement. Subsequent approaches, such as Self-Adaptive MAS~\citep{nascimento2023self}, further improve performance by enabling adaptive agent behaviors and dynamic coordination strategies. More recent work~\citep{dang2025multi} advances this direction by training a dedicated controller (or “puppeteer”) model to manage agent selection and pipeline execution, thus constantly improving orchestration efficiency and overall performance of cross domain tasks. 

\textbf{CoT and RL}
Chain-of-Thought (CoT) reasoning and reinforcement learning (RL) have demonstrated 
strong effectiveness in improving multi step reasoning and decision making in recent 
studies~\citep{guo2025deepseek, wei2022chain, singh2025agentic, yan2025memory}. These techniques 
have been increasingly adopted across diverse domains to enhance reasoning quality, 
planning depth, and task robustness. For example, KBQA-o1~\citep{luo2025kbqa} and Search-R1~\citep{jin2025search} leverage 
agent based reasoning and tool augmented query strategies to improve complex question 
answering and search oriented tasks. These approaches integrate stepwise reasoning with 
environment interaction, allowing models to iteratively retrieve evidence, revise plans, 
and verify intermediate conclusions. Beyond general reasoning benchmarks, CoT and 
RL driven frameworks have also been applied to specialized domains, including 
legal reasoning~\citep{cai-etal-2025-unilaw}, mathematical problem solving~\citep{huan2025does}, 
and financial decision systems~\citep{xiao2025trading}.  More recent work further combines CoT style reasoning with policy optimization methods~\citep{huang2025mobilevla, wei2025webagent}. As a result, 
CoT+RL pipelines are increasingly viewed as a practical foundation.

\begin{table*}[!t]
\centering
\caption{Emergent role taxonomy and action space in Our framework. 
Roles are not predefined; the policy model learns to generate 
Agent Roles strings via RL. We categorize observed emergent roles by their functional stage in multi turn reasoning.}
\label{tab:emergent_roles}
\resizebox{\textwidth}{!}{
\begin{tabular}{@{}llllll@{}}
\toprule
\textbf{Category} & \textbf{Emergent Role} $r \in \mathcal{R}$ 
  & \textbf{Typical Output} & \textbf{Stage} 
  & \textbf{Action} & \textbf{State Update} \\
\midrule
Information Gathering 
  & \texttt{Researcher}, \texttt{Domain Expert} 
  & background facts, evidence 
  & Early 
  & \texttt{tool\_call}$(r, p)$ 
  & $\mathcal{H}_t \leftarrow \mathcal{H}_{t-1} \oplus (p, y_r)$ \\
Planning 
  & \texttt{Planner}, \texttt{Strategist} 
  & strategy, subtask decomposition 
  & Early 
  & \texttt{tool\_call}$(r, p)$ 
  & $\mathcal{H}_t \leftarrow \mathcal{H}_{t-1} \oplus (p, y_r)$ \\
Domain Solving 
  & \texttt{Math Solver}, \texttt{Coder}, \texttt{Psychologist} 
  & solution, code, domain analysis 
  & Middle 
  & \texttt{tool\_call}$(r, p)$ 
  & $\mathcal{H}_t \leftarrow \mathcal{H}_{t-1} \oplus (p, y_r)$ \\
Verification 
  & \texttt{Verifier}, \texttt{Clarifier} 
  & validation, error identification 
  & Late 
  & \texttt{tool\_call}$(r, p)$ 
  & $\mathcal{H}_t \leftarrow \mathcal{H}_{t-1} \oplus (p, y_r)$ \\
Critique 
  & \texttt{Critiquer} 
  & evaluation score, improved content 
  & Late 
  & \texttt{tool\_call}$(r, p, \theta_c)$ 
  & $\mathcal{H}_t \leftarrow \mathcal{H}_{t-1} \oplus (p, y_r^{\text{crit}})$ \\
Synthesis 
  & \texttt{Assistant} (default) 
  & final answer $y_q$ 
  & Terminal 
  & \texttt{stop} 
  & $y_q = \pi_\phi(\mathcal{H}_T, q)$ \\
\bottomrule
\end{tabular}
}
\label{table:role-spec}
\end{table*}

\section{Methodology}
This section introduces our proposed approach OrchMAS, as shown in figure~\ref{fig:mas_page3}, an architecture featuring adaptive role orchestration in table~\ref{table:role-spec}, a layered critique refinement paradigm, and holistic RL driven by composite reward signals across multi stage reasoning trajectories and execution feedback.

\subsection{Adaptive MAS Coordination}
Within OrchMAS, heterogeneous specialized agents interact via adaptive role allocation and structured hierarchical management under dynamic controller guidance and runtime policy constraints. The complete operational workflow proceeds as follows:

\textbf{System Bootstrapping.}
OrchMAS employs a coordinator worker paradigm with flexible role instantiation. The MAS initialization comprises the following components:

\textit{(i) Interaction Substrate $\mathcal{E}$.}
The interaction substrate \(\mathcal{E}\) functions as the mediation layer within OrchMAS. The substrate \(\mathcal{E}\) dispatches communication requests \(u_t^{\mathrm{msg}}\) to designated agents according to role designation \(\sigma_t\) and dialogue history \(H_{t-1}\):
\begin{equation}
v_t^{\mathrm{msg}} \sim \mathbf{{\mathcal{E}}}\big(\cdot \mid H_{t-1}, u_t^{\mathrm{msg}}, \sigma_t\big).
\end{equation}
The feedback \(v_t^{\mathrm{msg}}\) from the interaction substrate \(\mathcal{E}\) together with the communication request \(u_t^{\mathrm{msg}}\) are encapsulated with boundary tokens to refresh the collaborative dialogue history \(H_t\):
\begin{equation}
H_t = H_{t-1} \oplus \big(\langle\mathrm{query}\rangle u_t^{\mathrm{msg}} \langle/\mathrm{query}\rangle, \langle\mathrm{reply}\rangle v_t^{\mathrm{msg}} \langle/\mathrm{reply}\rangle\big).
\end{equation}
where \(\oplus\) signifies concatenating the fresh query reply tuple with boundary tokens. At initialization, the history is \(H_0 = [\,]\) and the feedback is \(v_0^{\mathrm{msg}} = \varnothing\).

\textit{(ii) Coordinator Module \(\mathcal{C}\).}
The coordinator LLM \(\mathcal{C}\) manages the MAS ecosystem. At the outset, \(\mathcal{C}\) examines the query \(x\) and picks a suitable role \(\sigma_1\) from the role repertoire \(\mathcal{S}\) (see Table~\ref{table:role-spec}), producing the inaugural round of deliberation \(u_1^{\mathrm{reason}}\) and communication request $u_1^{\mathrm{msg}}$ with role designation:
\begin{equation}
\big(u_1^{\mathrm{reason}}, u_1^{\mathrm{msg}}, \sigma_1\big) \sim  \mathbf{{\mathcal{C}}}\big(\cdot \mid x, \mathcal{S}\big).
\end{equation}
Subsequently, the request $u_1^{\mathrm{msg}}$ with role $\sigma_1$ is relayed through $\mathcal{E}$, and the feedback $v_1^{\mathrm{msg}}$ constitutes the inaugural round history $H_{1}$ for ensuing interactions.

\textbf{Collaborative State Manifold $\mathcal{H}$.}
The coordinator state $h_t$ is characterized by the accumulated collaborative dialogue history of $\mathcal{C}$, including prior role assignments, agent feedback signals, and orchestration decisions across turns, forming a structured latent context for subsequent planning and control.

\textit{(i) Genesis State (\(h_0\)).}
The genesis state of $\mathcal{C}$ is \(h_0\), \(h_0 = [\,]\). The state of the inaugural round $h_1$ derives from \(x\) and role repertoire \(\mathcal{S}\), comprising $u_1^{\mathrm{reason}}$, $u_1^{\mathrm{msg}}$, $\sigma_1$, and $v_1^{\mathrm{msg}}$. The inaugural round state \(h_1\) manifests as:
\begin{equation}
h_1 = [x \oplus \mathcal{S} \oplus u_1^{\mathrm{reason}} \oplus (\sigma_1, u_1^{\mathrm{msg}}) \oplus v_1^{\mathrm{msg}}]
\end{equation}

\textit{(ii) State Evolution (\(h_t\)).}
Beginning from round 2, the state evolution of $\mathcal{C}$ hinges on $h_{t-1}$, integrating role guided deliberation $u_t^{\mathrm{reason}}$, request $u_t^{\mathrm{msg}}$ with role $\sigma_t$, and feedback $v_t^{\mathrm{msg}}$:

\begin{equation}
h_{t}=[u_t^{\mathrm{reason}} \oplus (\sigma_t, u_t^{\mathrm{msg}}) \oplus v_t^{\mathrm{msg}}].
\end{equation}

\textit{(iii) History Encoding (\(G_{h_t}\)).}
The history encoding \( G_{h_t} \) aggregates the entire MAS interaction chronicle up to round \( t \), capturing all deliberations, role guided requests, and feedbacks. At round \( t \), the encoding \( G_{h_t} \) is refreshed by fusing the antecedent encoding \( G_{h_{t-1}} \) with the current round's elements:

\begin{equation}
G_{h_t} = {\mathbf{\mathcal{C}}_{t}}(G_{h_{t-1}}, u_t^{\text{reason}}, \sigma_t, u_t^{\text{msg}}, v_t^{\text{msg}} ),
\end{equation}

\textbf{Collaborative Action Manifold.}
The coordinator \(\mathcal{C}\) determines subsequent actions encompassing role selection and request construction until the halting criterion is fulfilled:

\begin{equation}
\begin{aligned}
\log \mathbf{{\phi_\omega}}\big(u_t \mid G_{h_t}\big) &= \log \mathbf{{\phi_\omega}}\big(u_t^{\mathrm{reason}} \mid G_{h_{t-1}}\big) \\
& \hspace{-1.6cm} + \log \mathbf{{\phi_\omega}}\big(\sigma_t, u_t^{\mathrm{msg}} \mid G_{h_{t-1}}, u_t^{\mathrm{reason}}\big).
\end{aligned}
\end{equation}
Throughout this procedure, the coordinator $\mathcal{C}$ assesses role guided action likelihoods under a stochastic strategy, guiding the MAS execution path toward the ultimate solution.

\textbf{Coordinator Objective \((h_\ell, G_{h_\ell}, Y_{h_\ell})\).}
Following MAS cooperation through the substrate $\mathcal{E}$, the coordinator \(\mathcal{C}\) assembles the definitive solution to query \(x\).

\textit{(i) Terminus State:}

The cooperation concludes at round \(T\), where the interaction process reaches a stable termination condition and no further role reallocation or pipeline revision is triggered. The terminus encoding \(G_{h_T}\) is grounded in the complete dialogue chronicle \(H_T\), which aggregates all intermediate reasoning traces, coordination actions, and the concluding feedback \(v_T^{\mathrm{msg}}\). This finalized collaborative context is furnished to $\mathcal{C}$ to produce the definitive and globally consolidated solution $Y_{h_\ell}$:
\begin{equation}
z = \arg\max_{z \in \mathcal{V}^\ast} \mathbf{{\phi_\omega}}\big(z \mid x, \mathcal{S}, H_T\big),
\end{equation}
where \(z\) represents the solution \(Y_{h_\ell}\) assembled by the coordinator \(\mathcal{C}\) under full-history conditioning.

\textit{(ii) Combined Distribution:}
The combined distribution of the MAS coordination procedure for the coordinator \(\mathcal{C}\) and the substrate \(\mathcal{E}\) is expressed as:
\begin{equation}
{\fontsize{9.5pt}{10pt}\selectfont
\begin{aligned}
&Q_\omega(\xi, z \mid x, \mathcal{S}) = \\
&\underbrace{\mathbf{{\phi_\omega}}(u^{\mathrm{reason}}_1, \sigma_1, u^{\mathrm{msg}}_1 \mid x, \mathcal{S})}_{\text{Inaugural round: Role guided request}} \,
\underbrace{\mathbf{{\mathcal{E}}}(v^{\mathrm{msg}}_1 \mid H_0, \sigma_1, u^{\mathrm{msg}}_1)}_{\text{Inaugural round: Agent feedback}} \\
&\times \prod_{t=2}^{T} \big( \underbrace{\mathbf{{\phi_\omega}}(u^{\mathrm{reason}}_t, \sigma_t, u^{\mathrm{msg}}_t \mid x, \mathcal{S}, H_{t-1})}_{\text{Following rounds: Role guided request}} \big) \\
&\times \big( \underbrace{\mathbf{{\mathcal{E}}}(v^{\mathrm{msg}}_t \mid H_{t-1}, \sigma_t, u^{\mathrm{msg}}_t)}_{\text{Following rounds: Agent feedback}} \big) \\
&\times \underbrace{\mathbf{{\phi_\omega^{\mathrm{sol}}}}(u^{\mathrm{reason}}_t, z \mid x, \mathcal{S}, H_T)}_{\text{Conclusive assembly}}.
\end{aligned}}
\end{equation}
where $\xi=\big\{(x, \mathcal{S}, \sigma_t, u^{\mathrm{msg}}_t, v^{\mathrm{msg}}_t)\big\}_{t=1}^{T}$ denotes the execution path for MAS coordination; $\phi_\omega$ represents the coordination strategy of $\mathcal{C}$; and $Q_\mathcal{E}$ is the conditional distribution of the interaction substrate $\mathcal{E}$ routing to role designated agents.

\begin{table*}[t]
\centering
\caption{Our two level prompt architecture. \textbf{(a)} The policy model $\pi_\phi$ follows a multi turn interaction protocol: \textcolor{orange}{\textbf{think}} $\to$ \textcolor{green!50!black}{\textbf{interact}} (with \textcolor{red!80!black}{\textbf{agent\_role}}) $\to$ \textcolor{cyan!60!black}{\textbf{observe}} $\to$ $\cdots$ $\to$ \textcolor{blue}{\textbf{answer}}. \textbf{(b)} The external LLM $\mathcal{M}_{\text{ext}}$ receives a dynamically constructed system prompt where \textcolor{red!80!black}{\textbf{\texttt{agent\_role}}}, learned entirely through RL, determines its persona and expertise.}
\small
\renewcommand{\arraystretch}{0.9}
\begin{tabular}{p{1\textwidth}}
\toprule
\multicolumn{1}{c}{\textbf{(a) Policy Model Prompt $\pi_\phi$ --- Interaction Protocol}} \\
\midrule
\textcolor{purple}{\textbf{Question}} \\[2pt]
First, provide a simple explanation of the question and give it to the large language model for a more accurate answer. Focus on explaining the question without deep reasoning in the first step. After receiving the response, think about the large language model's response, and by interacting with the large language model again and again, arrive at the final answer. Proceed step by step with the following rules: \\[2pt]
1. Only in the first step, provide a brief explanation of the question and give it to the large language model: \\
\quad \textcolor{orange}{\textbf{\texttt{<think>}}}(Brief thinking must not be over 80 words)\textcolor{orange}{\textbf{\texttt{</think>}}} \\
\quad \textcolor{green!50!black}{\textbf{\texttt{<interaction\_prompt>}}}\texttt{\{"name": "prompt\_dynamic", "arguments": \{"prompt": "...", }\textcolor{red!80!black}{\textbf{"agent\_role": "\{role\}"}}\texttt{\}\}}\textcolor{green!50!black}{\textbf{\texttt{</interaction\_prompt>}}} \\[2pt]
2. After the first step, in each interaction with the large language model, write: \\
\quad \textcolor{orange}{\textbf{\texttt{<think>}}}(your reasoning for the receiving response and question)\textcolor{orange}{\textbf{\texttt{</think>}}} \\
\quad \textcolor{green!50!black}{\textbf{\texttt{<interaction\_prompt>}}}(new request with \textcolor{red!80!black}{\textbf{agent\_role}} to refine or validate the answer)\textcolor{green!50!black}{\textbf{\texttt{</interaction\_prompt>}}} \\[2pt]
3. Each \textcolor{green!50!black}{\textbf{\texttt{<interaction\_prompt>}}} must build on what came before. Do not just repeat the same content. Let the content evolve naturally (for example: outline $\to$ add details $\to$ refine $\to$ check). \\[2pt]
4. Continue producing think within \textcolor{orange}{\textbf{\texttt{<think>...</think>}}} and call tool within \textcolor{green!50!black}{\textbf{\texttt{<interaction\_prompt>...<\!/interaction\_prompt>}}} until the answer is ready. \\[2pt]
5. Once the answer is complete, write: \\
\quad \textcolor{orange}{\textbf{\texttt{<think>}}}(final reasoning with the \textcolor{cyan!60!black}{\textbf{\texttt{<interaction\_response>}}} and question)\textcolor{orange}{\textbf{\texttt{</think>}}} \\
\quad \textcolor{blue}{\textbf{\texttt{<answer>}}}(final answer for the question)\textcolor{blue}{\textbf{\texttt{</answer>}}} \\
\midrule
\multicolumn{1}{c}{\textbf{(b) External LLM Prompt $\mathcal{M}_{\text{ext}}$ --- Dynamic Role Assignment}} \\
\midrule
\textbf{System:} \; ``You are \textcolor{red!80!black}{\textbf{\{agent\_role\}}}. Please read the provided content (including previous conversations and the current task) and help the user complete the task or answer the question.'' \\[4pt]
\textbf{History:} \; $\mathcal{H}_{<t} = [(p_1, y_1), \ldots, (p_{t-1}, y_{t-1})]$ \quad {\scriptsize (prior multi turn conversation messages)} \\[4pt]
\textbf{User:} \; \textcolor{green!50!black}{\textbf{\{prompt\}}} \quad {\scriptsize (the task/question composed by $\pi_\phi$)} \\[4pt]
\textcolor{red!80!black}{\textbf{\texttt{agent\_role}}} is a \emph{free form string} generated by $\pi_\phi$ via RL---not selected from a predefined set. \\
Observed emergent roles include: \texttt{Researcher}, \texttt{Planner}, \texttt{Math Solver}, \texttt{Coder}, \texttt{Psychologist}, \texttt{Verifier}, \texttt{Clarifier}, \texttt{Assistant}, etc. \\
\bottomrule
\end{tabular}
\renewcommand{\arraystretch}{1.0}
\label{tab:prompt}
\end{table*}

\subsection{Layered Critique Refinement Learning}

We refine the coordinator's strategy \( \phi_\omega \) via a layered reward integrating structural soundness and solution precision, with holistic RL (prompts shown in table~\ref{tab:prompt}). The reward \( \mathcal{R} \) encapsulates Format adherence and the response fidelity at the execution path level.

\textbf{Layered Reward Architecture.}
To enforce MAS cooperation and accurate solutions, we define two layered rewards $\mathcal{C}$ output: the \emph{Format reward} $\mathcal{R}_{\text{fmt}}$ and the \emph{precision reward} $\mathcal{R}_{\text{prec}}$.

\textit{(i) Format Reward.}
At round $t$, both deliberation and role-guided requests must be well structured:
$\Lambda_t=\mathbb{I}[u_t^{\mathrm{reason}}\neq\varnothing \wedge (\sigma_t, u_t^{\mathrm{msg}})\neq\varnothing]$.
At the final turn, we additionally require valid boundary tokens $B_m, B_v, I_c$ to ensure structural correctness:
\begin{equation}
\mathcal{R}_{\mathrm{fmt}}=\!\big(\kappa,\;\alpha\!\sum_{t=1}^{T-1}\! \Lambda_t + \beta B_m + \gamma B_v + \delta I_c\big)
\end{equation}
where \( \Lambda_t \) verifies properly structured deliberation and role-guided requests, preventing malformed cooperation; \( B_m \) checks boundary token validity for query–reply encapsulation; \( B_v \) ensures interpretability of the final assembly; \( I_c \) enforces structural completeness; coefficients \( (\alpha, \beta, \gamma, \delta) \) balance intermediate cooperation and valid assembly objectives; and \( \kappa \) caps the Format reward to stabilize optimization dynamics and training behavior.

\emph{(ii) Precision Reward.}
Let $\hat{s}=\mathrm{Std}(\mathrm{Asm}(z))$ be the standardized assembled solution by the coordinator $\mathcal{C}$, and
$\mathcal{T}(x)=\{t_i\}$ the reference collection. Standardization
$\mathrm{Std}(\cdot)$ eliminates case, punctuation, and articles, while
$\mathrm{Seg}(\cdot)$ transforms text to a multiset of segments.
The segment level F1 with a reference $t$ is specified as:
computed over segment overlap precision and recall statistics.
\begin{equation}
\mathrm{F1}(\hat s, t)=\frac{2\,n_{\cap}}{\,|\mathrm{Seg}(\hat s)|+|\mathrm{Seg}(\mathrm{Std}(t))|\,}.
\end{equation}
where $n_{\cap}$ is the segment overlap count between the assembled solution and ground truth.
The correctness of the assembled solution (precision reward) is:
\begin{equation}
\mathcal{R}_{\text{prec}} = \max_{t \in \mathcal{T}(x)} \mathrm{F1}(\hat{s}, t).
\end{equation}
\emph{(iii) Gated Aggregation for Layered Reward.}
The aggregate reward $\mathcal{R}$ incorporates the Format reward $\mathcal{R}_{\text{fmt}}$ and the solution precision reward $\mathcal{R}_{\text{prec}}$. The computation of the aggregate reward $\mathcal{R}$ is:
\begin{equation}
\mathcal{R} =
\begin{cases}
-\kappa + \mathcal{R}_{\text{fmt}} + \mathcal{R}_{\text{prec}}, & \mathcal{R}_{\text{fmt}}=\kappa,\\
-\kappa + \mathcal{R}_{\text{fmt}}, & \text{otherwise}.
\end{cases}
\end{equation}
so that the precision of the assembly is only acknowledged when the Format prerequisites are completely and correctly fulfilled, ensuring strict structural validity before accuracy based reward activation and preventing premature scoring under partially valid or structurally inconsistent outputs, thereby improving training stability and discouraging shortcut style reward exploitation.

\textbf{Holistic RL.}
We adopt a GRPO style objective, normalizing rewards across a batch of \(M\) execution paths. Let \(\mathcal{R}^{(i)}\) be the reward of execution path \(i\), with the mean reward \(\bar{\mathcal{R}}\):
\begin{equation}
\hat{W}^{(i)} = \frac{\mathcal{R}^{(i)} - \bar{\mathcal{R}}}{\sqrt{\tfrac{1}{M}\sum_{j=1}^M (\mathcal{R}^{(j)}-\bar{\mathcal{R}})^2 + \varepsilon}},
\end{equation}
where $\hat{W}^{(i)}$ is the normalized advantage, and $\varepsilon$ is a stabilization constant.
The A-GRPO (Agent GRPO) style objective is:
\begin{equation}
{\fontsize{10pt}{10pt}\selectfont
\begin{aligned}
\mathcal{L}_{\text{A-GRPO}}(\omega) &= \mathbb{E}_{\xi \sim q_{{\omega_\mathcal{C}}, {\omega_\mathcal{E}}}(\xi)} \big[ \frac{1}{M} \sum_{i=1}^M \big( \frac{1}{|\xi^{(i)}|}\sum_{t=1}^{|\xi^{(i)}|} \min \big( \\
&\hspace{-1.3cm} \frac{{\phi_\omega}(w_t^{(i)} \mid \xi_{<t}^{(i)})}{{\phi_{\omega_{\text{prev}}}}(w_t^{(i)} \mid \xi_{<t}^{(i)})} \hat{W}(\xi^{(i)}), \text{clip}\!\big(\frac{{\phi_\omega}(w_t^{(i)} \mid \xi_{<t}^{(i)})}{\phi_{\omega_{\text{prev}}}(w_t^{(i)} \mid \xi_{<t}^{(i)})},1 \pm \epsilon \big)\\
& \hspace{-1.3cm} \times \hat{W}(\xi^{(i)}) \big) - \lambda \mathbb{D}_{\text{KL}}({\phi_\omega} \parallel \phi_{\text{base}}) \big) \big],
\end{aligned}}
\end{equation}
where $q_{\omega_\mathcal{C}, \omega_\mathcal{E}}(\xi)$ is the combined distribution of coordinator $\mathcal{C}$ and substrate $\mathcal{E}$; $w_t^{(i)}$ is the $t$-th token of $\xi^{(i)}$, and $\phi_{\omega_{\text{prev}}}$; and $\phi_{\text{base}}$ are the pre update and baseline strategies, respectively. The $\text{clip}(\cdot)$ constrains strategy ratios to $1 \pm \epsilon$ to stabilize updates. A KL term $\mathbb{D}_{\text{KL}}({\phi_\omega} \parallel \phi_{\text{base}})$ regularizes the baseline strategy \( \phi_{\text{base}} \) with \( \lambda \) governing its magnitude.

\begin{table*}[!t]
\small
\caption{
performance comparison across six QA benchmarks (2Wiki, HotpotQA, GSM8K, DAPO, MusiQue, and PopQA) measured by F1 and Exact Match (EM). We compare baseline models, supervised fine tuning (SFT), CoT prompting, GRPO training, and representative MAS optimization methods (OPRO, TextGrad, GEPA) against our proposed OrchMAS framework. Our method consistently achieves the best results across all datasets, with substantial gains over the strongest non OrchMAS baseline.
}
\begin{tabular}{cccccccccccccc}
\hline
\multicolumn{2}{c}{Methods} &
  \multicolumn{2}{c}{\textbf{2Wiki}} &
  \multicolumn{2}{c}{\textbf{Hotpot}} &
  \multicolumn{2}{c}{\textbf{GSM8K}} &
  \multicolumn{2}{c}{\textbf{DAPO}} &
  \multicolumn{2}{c}{\textbf{PopQA}} &
  \multicolumn{2}{c}{\textbf{MusiQue}} \\ \cline{3-14} 
\multicolumn{2}{c}{\textbf{}}                    & F1      & EM      & F1      & EM      & F1     & EM     & F1      & EM      & F1     & EM     & F1     & EM     \\ \hline
\multirow{2}{*}{Baseline}          & Qwen3-4B    & 28.21   & 26.04   & 23.21   & 21.88   & 84.38  & 84.38  & 0.00    & 0.00    & 11.02  & 7.29   & 6.71   & 3.13   \\
                                   & GPT-4o-mini & 35.44   & 33.33   & 41.21   & 32.29   & 86.61  & 83.33  & 19.79   & 19.79   & 32.30  & 25.00  & 23.59  & 14.58  \\ \hline
SFT                                & Qwen3-4B    & 38.91   & 41.67   & 33.09   & 23.96   & 32.29  & 32.29  & 5.21    & 5.21    & 8.89   & 9.38   & 18.61  & 8.33   \\ \hline
\multirow{2}{*}{CoT Reasoning}     & Qwen3-4B    & 25.21   & 20.83   & 24.98   & 18.75   & 81.25  & 81.25  & 0.00    & 0.00    & 10.97  & 7.29   & 8.72   & 4.17   \\
                                   & GPT-4o-mini & 50.89   & 43.75   & 46.67   & 40.63   & 87.02  & 84.38  & 22.92   & 22.92   & 33.92  & 28.13  & 26.33  & 17.71  \\ \hline
GRPO                               & Qwen3-4B    & 34.02   & 35.42   & 31.21   & 23.96   & 92.71  & 92.71  & 6.25    & 6.25    & 14.48  & 10.41  & 12.98  & 9.38   \\ \hline
\multirow{3}{*}{MAS (GPT-4o-mini)} & OPRO        & 34.36   & 26.04   & 41.82   & 36.45   & 82.11  & 63.54  & 7.31    & 7.29    & 31.12  & 26.04  & 25.11  & 14.58  \\
                                   & TextGrad    & 26.42   & 17.71   & 35.21   & 28.13   & 81.21  & 70.83  & 12.51   & 11.46   & 29.32  & 19.79  & 22.06  & 15.63  \\
                                   & GEPA        & 41.24   & 37.50   & 47.13   & 39.58   & 91.10  & 84.38  & 15.16   & 13.54   & 32.15  & 27.08  & 23.21  & 15.63  \\ \hline
\multirow{2}{*}{Ours} &
  OrchMAS &
  \textbf{67.25} &
  \textbf{60.42} &
  \textbf{61.99} &
  \textbf{53.13} &
  \textbf{97.92} &
  \textbf{97.92} &
  \textbf{56.64} &
  \textbf{56.64} &
  \textbf{39.87} &
  \textbf{34.38} &
  \textbf{33.33} &
  \textbf{21.88} \\
                                   & delta       & + 16.36 & + 16.67 & + 14.86 & + 12.50 & + 5.21 & + 5.21 & + 33.72 & + 33.72 & + 5.95 & + 6.25 & + 7.00 & + 4.17 \\ \hline
\end{tabular}
\label{tab:main_results}
\end{table*}
\section{Experiments}
This section presents the experimental setup, results, and analysis. We study the following research questions (RQs): \textbf{RQ1}: Does our MAS framework outperform existing MAS LLM systems in overall QA and reasoning performance? \textbf{RQ2}: Can our framework achieve stronger and more consistent results across diverse task types and benchmarks? \textbf{RQ3}: Is the framework more effective and robust on long context and long text QA tasks? \textbf{RQ4}: Can our framework gain improvements on OOD tasks? \textbf{RQ5}: What is the contribution of each major framework component to the final performance according to ablation analysis? 

\subsection{Experimental Setup}
\textbf{Datasets. }
We evaluate our framework on a diverse set of QA and summarization benchmarks that approximate scientific and knowledge intensive problem settings, including multi hop reasoning, numerical reasoning, evidence aggregation, fact verification, and long form analytical summarization. 
These tasks require structured decomposition, cross source evidence integration, and multi step logical inference, which are core characteristics of scientific reasoning workflows. 
Our in distribution training and evaluation datasets include 2WikiMultiHopQA (2Wiki)~\citep{ho2020constructing}, HotpotQA~\citep{yang-etal-2018-hotpotqa}, GSM8K~\citep{cobbe2021training}, DAPO~\citep{yu2025dapo}, MusiQue~\citep{trivedi2022musique}, PopQA~\citep{mallen-etal-2023-trust}, BookSum~\citep{kryscinski-etal-2022-booksum}, and WritingPrompts (W.P.)~\citep{huang2024gpt}. 
These benchmarks collectively stress multi step deduction, quantitative reasoning, knowledge grounding, and long context synthesis, resembling core capabilities required in scientific and technical domains. 
To test OOD generalization under domain and task shifts, we further evaluate on TriviaQA~\citep{joshi2017triviaqa}, MathQA~\citep{amini-etal-2019-mathqa}, SQuAD v2~\citep{rajpurkar2018know}, and XSum.

\textbf{Baselines. }
We compare against several representative baselines: Direct prompting baselines using Qwen3-4B and GPT-4o-mini; SFT on the base model; CoT prompting; GRPO based RL on Qwen3-4B; and prior other MAS optimization approaches, including OPRO~\citep{yang2024llm_optimizers}, TextGrad~\citep{yuksekgonul2024textgrad}, and GEPA~\citep{agrawal2025gepa}, implemented with GPT-4o-mini.

\textbf{Evaluation Metrics. }
We report Exact Match (EM) and token level F1, but for summarization, we report cosine similarity, as it better reflects semantic alignment and content level agreement beyond strict lexical overlap.

\textbf{Hyperparameters and Environment. }
The orchestrator is based on Qwen3-4B trained with GRPO, while the executor is a locally deployed GPT-OSS-120B model. Training and inference are conducted on two separate 4$\times$RTX A6000 GPU nodes, one for GRPO training and one for large-model inference. We use a maximum prompt length of 8192 tokens and limit each interaction episode to at most 5 agent turns. 
Key GRPO settings include a learning rate of $1\times10^{-6}$, KL regularization coefficient 0.001, PPO mini batch size 64, and repeated rollout sampling for stable reward estimation. 
\begin{table*}[!t]
\caption{
Performance comparison on in distribution and OOD tasks across summarization and question answering benchmarks. In distribution results are reported on BookSum and WritingPrompts (W.P.), while OOD generalization is evaluated on XSum, TriviaQA, MathQA, and SQuAD v2. 
}
\label{tab:long_context_results}
\centering
\begin{tabular}{cccc|ccccccc}
\hline
\multicolumn{2}{c}{\multirow{3}{*}{Methods}} &
  \multicolumn{2}{c|}{In Distribution} &
  \multicolumn{7}{c}{OOD Tasks} \\ \cline{3-11} 
\multicolumn{2}{c}{} &
  BookSum &
  W.P. &
  XSum &
  \multicolumn{2}{c}{TriviaQA} &
  \multicolumn{2}{c}{MathQA} &
  \multicolumn{2}{c}{SQuAD v2} \\ \cline{3-11} 
\multicolumn{2}{c}{}                             & Cos   & Cos   & Cos   & F1      & EM     & F1     & EM      & F1      & EM     \\ \hline
\multirow{2}{*}{Baseline}          & Qwen3-4B    & 42.32  & 21.55  & 32.64  & 46.54   & 44.79  & 31.86  & 29.17   & 16.22   & 7.29   \\
                                   & GPT-4o-mini & 57.43  & 33.37  & 61.33  & 76.33   & 63.54  & 55.29  & 46.88   & 25.41   & 13.54  \\ \hline
SFT                                & Qwen3-4B    & 52.87  & 29.68  & 54.61  & 32.82   & 31.25  & 23.74  & 17.71   & 16.14   & 5.21   \\ \hline
\multirow{2}{*}{CoT Reasoning}     & Qwen3-4B    & 35.32  & 14.13  & 6.37   & 51.77   & 46.88  & 32.13  & 27.08   & 16.98   & 6.25   \\
                                   & GPT-4o-mini & 52.17  & 32.42  & 51.71  & 74.79   & 67.71  & 56.51  & 48.96   & 24.37   & 15.63  \\ \hline
GRPO                               & Qwen3-4B    & 51.17  & 11.31  & 52.33  & 56.93   & 51.04  & 53.38  & 45.83   & 25.28   & 10.42  \\ \hline
\multirow{3}{*}{MAS (GPT-4o-mini)} & OPRO        & 24.38  & 15.04  & 28.14  & 71.24   & 60.41  & 59.92  & 44.79   & 24.56   & 11.46  \\
                                   & TextGrad    & 33.09  & 22.21  & 25.91  & 68.51   & 54.17  & 62.36  & 44.79   & 21.07   & 7.29   \\
                                   & GEPA        & 0.00   & 0.46   & 0.42   & 71.67   & 66.67  & 68.23  & 40.63   & 23.68   & 13.54  \\ \hline
\multirow{2}{*}{Ours}              & OrchMAS        & 59.09  & 36.42  & 65.48  & 87.81   & 77.08  & 86.63  & 72.92   & 36.82   & 19.79  \\
                                   & delta       & + 1.66 & + 3.05 & + 4.15 & + 11.48 & + 9.37 & + 18.4 & + 23.96 & + 11.41 & + 4.16 \\ \hline
\end{tabular}
\end{table*}

\begin{figure}[!t]
\centering
\includegraphics[width=\linewidth]{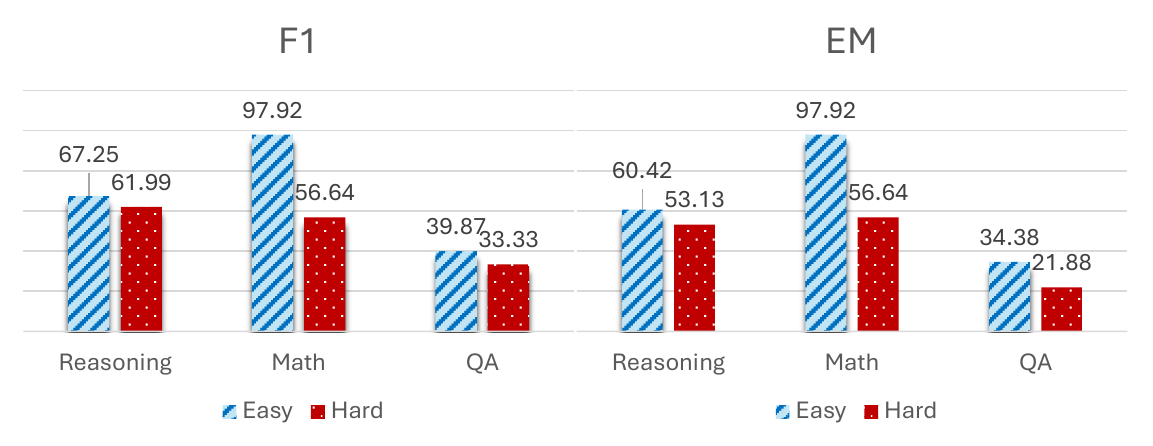}
\caption{Result comparison for easy and hard questions}
\label{compare}
\end{figure}
\subsection{OrchMAS's performance (RQ1)}
Table~\ref{tab:main_results} shows that our MAS framework (OrchMAS) consistently outperforms all compared methods, including representative MAS optimization approaches (OPRO, TextGrad, and GEPA), across all six QA benchmarks in both F1 and EM. The gains are stable across multi hop reasoning (2Wiki, HotpotQA, MusiQue), numerical reasoning (GSM8K), prompt optimization settings (DAPO), and open domain QA (PopQA), indicating strong cross task generalization. Compared with the strongest MAS baseline (GEPA), OrchMAS achieves large absolute improvements on complex multi hop datasets. On 2Wiki, OrchMAS improves F1 from 41.24 to 67.25 and EM from 37.50 to 60.42. On HotpotQA, F1 increases from 47.13 to 61.99 and EM from 39.58 to 53.13. Similar margins are observed on MusiQue and PopQA, where OrchMAS yields the highest scores among all systems. Notably, on DAPO, which is particularly sensitive to prompt and reasoning strategy quality, OrchMAS shows a substantial jump (F1/EM 56.64 vs. 15.16/13.54 for GEPA), suggesting that dynamic role and pipeline orchestration provides a stronger optimization signal than gradient style or prompt search MAS methods. We also observe that existing MAS optimization methods improve over vanilla baselines only modestly and inconsistently, whereas OrchMAS delivers uniform improvements across all benchmarks. The overall gains over the best non OrchMAS baseline average +16.36 F1 / +16.67 EM on 2Wiki, +14.86 / +12.50 on HotpotQA, and +33.72 / +33.72 on DAPO. 

These results support RQ1: our MAS framework surpasses existing MAS LLM systems in overall QA and reasoning performance. We attribute this advantage to dynamic task conditioned role generation and iterative pipeline adaptation, which reduce prompt–task mismatch and enable more effective multi step coordination than static role or fixed pipeline MAS designs.

\subsection{Various Task performance (RQ2)}
To systematically evaluate cross task robustness, we construct our benchmark suite to cover multiple task types and difficulty levels. Specifically, we select three representative categories: multi hop reasoning, mathematical problem solving, and open domain question answering. For each category, we include both a relatively easier dataset and a more challenging one. Concretely, 2Wiki (easy) and HotpotQA (hard) are used for reasoning, GSM8K (easy) and DAPO (hard) for math, and PopQA (easy) and MusiQue (hard) for QA. This design enables controlled comparison of model behavior across both task diversity and difficulty variation. Results in Table~\ref{tab:main_results} and Fig.~\ref{compare} show that OrchMAS consistently outperforms all baselines and competing MAS optimization methods across all six benchmarks. On reasoning tasks, OrchMAS achieves large gains on both 2Wiki and HotpotQA, indicating improved multi hop reasoning capability under both moderate and high compositional complexity. On mathematical tasks, OrchMAS reaches near ceiling performance on GSM8K and delivers especially large improvements on the more difficult DAPO benchmark, suggesting stronger stability in long step numerical reasoning. On QA tasks, OrchMAS also yields consistent gains on both PopQA and MusiQue, demonstrating that the benefits extend beyond reasoning and math into knowledge intensive question answering. Importantly, the improvements are not limited to a single task type or difficulty level, but appear consistently across easy–hard pairs in all three categories. This pattern indicates that OrchMAS improves general reasoning quality and orchestration effectiveness rather than overfitting to a specific benchmark format. Therefore, the results support that our framework achieves stronger and more consistent performance across diverse task types and benchmarks.
\begin{table*}[!t]
\caption{Ablation study of OrchMAS on diverse QA, reasoning, and summarization benchmarks. 
Results are reported using F1/EM for QA style tasks and Cos for summarization tasks. 
We compare the full OrchMAS framework with variants that remove one key module at a time: dynamic agent roles, multi turn reasoning, and environment guided execution. 
Removing any module leads to substantial and systematic performance degradation across datasets, demonstrating that each component plays a critical role and that their combination provides complementary gains in robustness, reasoning quality, and cross task generalization.
}
\label{tab:ablation}
\centering
\begin{tabular}{lccccccccccccc}
\hline
\multicolumn{2}{l}{\multirow{2}{*}{Dataset}} &
  \multicolumn{2}{c}{2Wiki} &
  \multicolumn{2}{c}{Hotpot} &
  \multicolumn{2}{c}{GSM8K} &
  \multicolumn{2}{c}{DAPO} &
  \multicolumn{2}{c}{MusiQue} &
  \multicolumn{2}{c}{PopQA} \\ \cline{3-14} 
\multicolumn{2}{l}{}                 & F1           & EM         & F1           & EM         & F1           & EM         & F1    & EM    & F1    & EM     & F1    & EM    \\ \hline
\multicolumn{2}{l}{OrchMAS}             & 67.25        & 60.42      & 61.99        & 53.13      & 97.92        & 97.92      & 56.64 & 56.64 & 33.33 & 21.88  & 39.87 & 34.38 \\
\multicolumn{1}{r}{} & - Agent Roles & 50.87        & 50.00      & 32.55        & 25.00      & 86.67        & 86.46      & 16.67 & 16.67 & 11.24 & 4.17   & 13.50 & 12.50 \\
\multicolumn{1}{r}{} & - Multi Turns & 38.43        & 37.5       & 53.30        & 44.79      & 78.13        & 78.13      & 42.71 & 42.71 & 9.50  & 5.21   & 39.39 & 35.42 \\
\multicolumn{1}{r}{} & - Environment & 45.00        & 43.75      & 36.36        & 28.13      & 96.88        & 96.88      & 16.67 & 16.67 & 13.59 & 6.25   & 18.51 & 15.63 \\ \hline
\multicolumn{2}{l}{\multirow{2}{*}{Dataset}} &
  \multicolumn{2}{c}{BookSum} &
  \multicolumn{2}{c}{W.P.} &
  \multicolumn{2}{c}{Xsum} &
  \multicolumn{2}{c}{Squad v2} &
  \multicolumn{2}{c}{TriviaQA} &
  \multicolumn{2}{c}{MathQA} \\ \cline{3-14} 
\multicolumn{2}{l}{}                 & \multicolumn{2}{c}{Cos}  & \multicolumn{2}{c}{Cos}  & \multicolumn{2}{c}{Cos}  & F1    & EM    & F1    & EM     & F1    & EM    \\ \hline
\multicolumn{2}{l}{OrchMAS}             & \multicolumn{2}{c}{59.09} & \multicolumn{2}{c}{36.42} & \multicolumn{2}{c}{65.48} & 36.82 & 19.79 & 87.81 & 77.08  & 86.63 & 72.92 \\
                     & - Agent Roles & \multicolumn{2}{c}{11.87} & \multicolumn{2}{c}{7.52}  & \multicolumn{2}{c}{10.35} & 22.64 & 9.38  & 57.72 & 53.13  & 64.41 & 54.17 \\
                     & - Multi Turns & \multicolumn{2}{c}{5.30}  & \multicolumn{2}{c}{7.94}  & \multicolumn{2}{c}{9.04}  & 30.76 & 16.67 & 86.34 & 78.125 & 78.82 & 67.71 \\
                     & - Environment & \multicolumn{2}{c}{11.52} & \multicolumn{2}{c}{8.06}  & \multicolumn{2}{c}{10.99} & 23.88 & 10.42 & 58.18 & 53.13  & 77.26 & 65.63 \\ \hline
\end{tabular}
\label{tab:compare}
\end{table*}
\begin{figure}[!t]
\centering
\includegraphics[width=\linewidth]{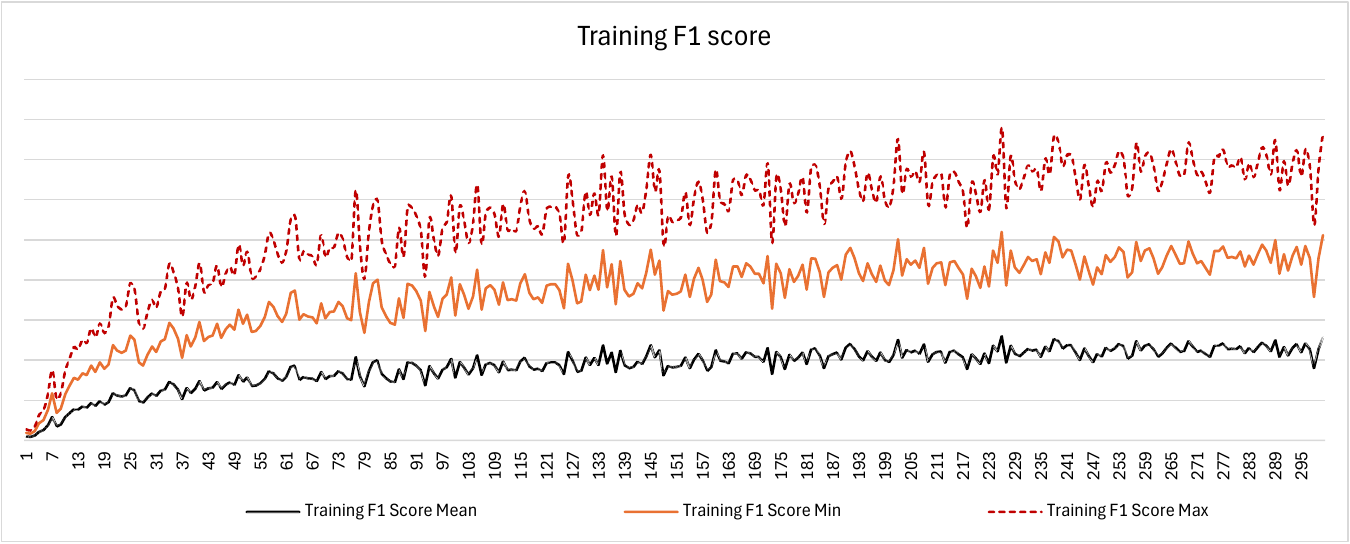}
\caption{Result comparison for easy and hard questions}
\label{fig:turn&lenth}
\end{figure}
\subsection{Long Context Performance (RQ3)}
To evaluate long context reasoning and compression capability, we conduct experiments on three summarization benchmarks, including BookSum and WritingPrompts, and XSum in table~\ref{tab:compare}. These tasks require models to process long inputs, identify salient information, and produce semantically faithful summaries. We report Cos scores to measure semantic similarity between generated summaries and references. From Table~\ref{tab:long_context_results}, OrchMAS consistently achieves the best performance across all summarization datasets. Compared with both backbone baselines and reasoning enhanced variants such as CoT and GRPO, OrchMAS yields clear gains on BookSum, WritingPrompts, and XSum. Notably, the improvements are observed not only on in distribution datasets but also on the OOD summarization benchmark, indicating stronger robustness under domain shift. We also observe that several optimization based or MAS prompt methods show unstable or degraded performance on long context summarization, suggesting that static prompting or shallow reasoning strategies are insufficient for long input compression. In contrast, OrchMAS’s dynamic MAS coordination and adaptive reasoning pipeline better preserves global context and key semantic structure, leading to higher quality summaries. These results support that OrchMAS provides more effective long context understanding capability.

\subsection{Performance On Unseen Tasks (RQ4)}
To evaluate cross task generalization, we test models on multiple OOD benchmarks that differ from the training setting in task format, reasoning pattern, and answer structure, including XSum summarization and three QA datasets: TriviaQA, MathQA, and SQuAD v2. These tasks cover factual retrieval, numerical reasoning, and extractive question answering, providing a diverse evaluation of transfer capability. As shown in Table~\ref{tab:long_context_results}, OrchMAS consistently outperforms all baselines and competing optimization or MAS prompt methods across OOD tasks. The gains are especially significant on reasoning intensive datasets such as MathQA and knowledge heavy datasets such as TriviaQA, where OrchMAS achieves large improvements in both F1 and EM. Figure~\ref{fig:turn&lenth} also showns its efficiency improvement. This indicates that the learned orchestration and dynamic role assignment strategy transfers beyond the original training distribution. In contrast, several strong baselines, including SFT and CoT prompting, show performance drops under distribution shift, and some prompt optimization MAS methods exhibit unstable behavior across tasks. OrchMAS maintains more stable performance across heterogeneous benchmarks.

\subsection{Contribution of Each Elements (RQ5)}
To analyze the contribution of each core component in OrchMAS, we conduct an ablation study across heterogeneous benchmarks, including multi hop QA, numerical reasoning, open domain QA, and long form summarization. We construct three ablated variants by removing (1) dynamic agent roles, (2) multi turn interaction, and (3) environment guided execution, while keeping model scale and training settings fixed for fair comparison. Results in Table~\ref{tab:ablation} show that removing any component consistently degrades performance across metrics (F1/EM and Cos), indicating that all modules provide generalizable gains. Removing dynamic roles causes the largest drops on knowledge intensive multi hop tasks, due to weaker task decomposition and role–task alignment. Removing multi turn interaction mainly hurts reasoning heavy datasets such as 2Wiki and GSM8K, highlighting the value of iterative refinement. Eliminating environment guided execution leads to broad declines, suggesting that structured feedback improves execution stability and reduces error propagation. Overall, the three components provide complementary benefits and jointly explain the robustness of our method.

\begin{figure}[!t]
\centering

\begin{minipage}[t]{0.48\linewidth}
\centering
\includegraphics[width=\linewidth]{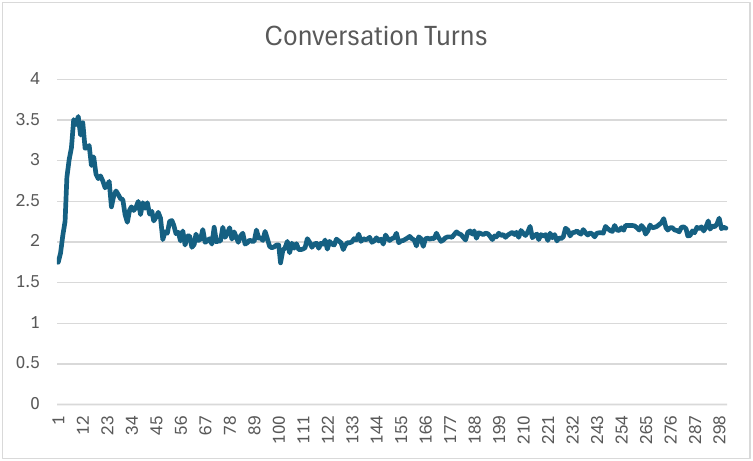}
\end{minipage}
\hfill
\begin{minipage}[t]{0.48\linewidth}
\centering
\includegraphics[width=\linewidth]{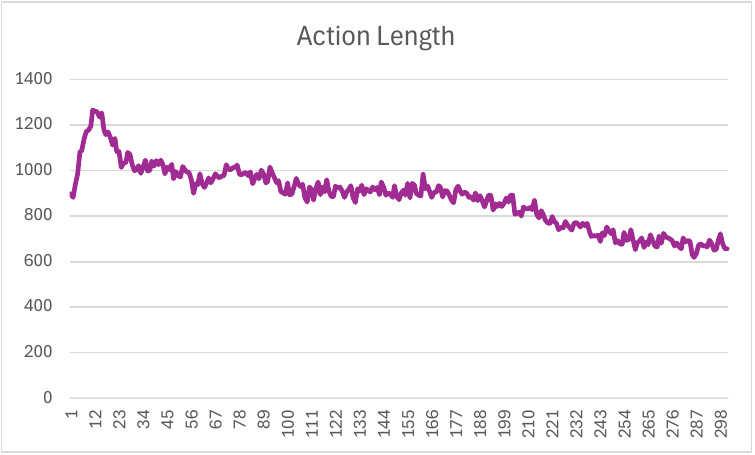}
\end{minipage}

\caption{Training time orchestration statistics.}
\label{fig:training_stats}

\end{figure}
\section{Conclusion}
In this work, we propose OrchMAS, a dynamic MAS orchestration framework for multi task and multi difficulty reasoning with improved adaptability and robustness, particularly for scientific and knowledge intensive problem solving. Unlike conventional MAS systems that rely on fixed prompts, static roles, and rigid pipelines, our framework introduces dynamic role assignment, iterative multi turn interaction, and guided execution for task aware reasoning and flexible coordination. This design alleviates prompt task misalignment and pipeline rigidity while improves the performances.

\section*{GenAI Disclosure}
We used Claude Code to assist with adding comments to the codebase, and ChatGPT-5.2 to help refine the writing and improve grammatical clarity in the manuscript.


\bibliographystyle{ACM-Reference-Format}
\bibliography{sample-base}

@inproceedings{wu2024affinity,
  title={On the affinity, rationality, and diversity of hierarchical topic modeling},
  author={Wu, Xiaobao and Pan, Fengjun and Nguyen, Thong and Feng, Yichao and Liu, Chaoqun and Nguyen, Cong-Duy and Luu, Anh Tuan},
  booktitle={Proceedings of the AAAI Conference on Artificial Intelligence},
  volume={38},
  number={17},
  pages={19261--19269},
  year={2024}
}

@article{althaf2025multi,
  title={Multi-Agent RAG Framework for Entity Resolution: Advancing Beyond Single-LLM Approaches with Specialized Agent Coordination},
  author={Althaf, Aatif Muhammad and Mohammed, Muzakkiruddin Ahmed and Milanova, Mariofanna and Talburt, John and Cakmak, Mert Can},
  journal={Computers},
  volume={14},
  number={12},
  pages={525},
  year={2025},
  publisher={MDPI}
}

@inproceedings{illusion-of-thinking,
title = {The Illusion of Thinking: Understanding the Strengths and Limitations of Reasoning Models via the Lens of Problem Complexity},
booktitle = {NeurIPS},
author = {Parshin Shojaee* and Iman Mirzadeh* and Keivan Alizadeh and Maxwell Horton and Samy Bengio and Mehrdad Farajtabar},
year = {2025},
URL = {https://arxiv.org/abs/2506.06941}
}

@article{li2024survey,
  title={A survey on LLM-based multi-agent systems: workflow, infrastructure, and challenges},
  author={Li, Xinyi and Wang, Sai and Zeng, Siqi and Wu, Yu and Yang, Yi},
  journal={Vicinagearth},
  volume={1},
  number={1},
  pages={9},
  year={2024},
  publisher={Springer}
}

@article{feng2025stimuli,
  title={From Stimuli to Minds: Enhancing Psychological Reasoning in LLMs via Bilateral Reinforcement Learning},
  author={Feng, Yichao and Luo, Haoran and Feng, Lang and Zhao, Shuai and Luu, Anh Tuan},
  journal={arXiv preprint arXiv:2508.02458},
  year={2025}
}

@article{feng2025aspect,
  title={Aspect-Based Summarization with Self-Aspect Retrieval Enhanced Generation},
  author={Feng, Yichao and Zhao, Shuai and Li, Yueqiu and Xiao, Luwei and Wu, Xiaobao and Luu, Anh Tuan},
  journal={arXiv preprint arXiv:2504.13054},
  year={2025}
}

@article{cemri2025multi,
  title={Why do multi-agent llm systems fail?},
  author={Cemri, Mert and Pan, Melissa Z and Yang, Shuyi and Agrawal, Lakshya A and Chopra, Bhavya and Tiwari, Rishabh and Keutzer, Kurt and Parameswaran, Aditya and Klein, Dan and Ramchandran, Kannan and others},
  journal={arXiv preprint arXiv:2503.13657},
  year={2025}
}

@article{perera2025auto,
  title={Auto-scaling LLM-based multi-agent systems through dynamic integration of agents},
  author={Perera, Ravindu and Basnayake, Anuradha and Wickramasinghe, Manjusri},
  journal={Frontiers in Artificial Intelligence},
  volume={8},
  pages={1638227},
  year={2025},
  publisher={Frontiers Media SA}
}

@article{talebirad2023multi,
  title={Multi-agent collaboration: Harnessing the power of intelligent llm agents},
  author={Talebirad, Yashar and Nadiri, Amirhossein},
  journal={arXiv preprint arXiv:2306.03314},
  year={2023}
}

@article{tran2025multi,
  title={Multi-agent collaboration mechanisms: A survey of llms},
  author={Tran, Khanh-Tung and Dao, Dung and Nguyen, Minh-Duong and Pham, Quoc-Viet and O'Sullivan, Barry and Nguyen, Hoang D},
  journal={arXiv preprint arXiv:2501.06322},
  year={2025}
}

@article{sun2025multi,
  title={Multi-agent coordination across diverse applications: A survey},
  author={Sun, Lijun and Yang, Yijun and Duan, Qiqi and Shi, Yuhui and Lyu, Chao and Chang, Yu-Cheng and Lin, Chin-Teng and Shen, Yang},
  journal={arXiv preprint arXiv:2502.14743},
  year={2025}
}

@article{xia2025parallelism,
  title={Parallelism Meets Adaptiveness: Scalable Documents Understanding in Multi-Agent LLM Systems},
  author={Xia, Chengxuan and Wu, Qianye and Tian, Sixuan and Hao, Yilun},
  journal={arXiv preprint arXiv:2507.17061},
  year={2025}
}

@article{su2025difficulty,
  title={Difficulty-Aware Agentic Orchestration for Query-Specific Multi-Agent Workflows},
  author={Su, Jinwei and Lan, Qizhen and Xia, Yinghui and Sun, Lifan and Tian, Weiyou and Shi, Tianyu and Song, Xinyuan and He, Lewei},
  journal={arXiv preprint arXiv:2509.11079},
  year={2025}
}

@article{chang2025sagallm,
  title={SagaLLM: Context Management, Validation, and Transaction Guarantees for Multi-Agent LLM Planning},
  author={Chang, Edward Y and Geng, Longling},
  journal={arXiv preprint arXiv:2503.11951},
  year={2025}
}

@article{lu2024morphagent,
  title={Morphagent: Empowering agents through self-evolving profiles and decentralized collaboration},
  author={Lu, Siyuan and Shao, Jiaqi and Luo, Bing and Lin, Tao},
  journal={arXiv preprint arXiv:2410.15048},
  year={2024}
}

@article{yediversity,
  title={Diversity for The Win: Towards Building Multi-Agent Systems with Heterogeneous LLMs},
  author={Ye, Rui and Liu, Xiangrui and Pang, Xianghe and Wu, Qimin and Yin, Zhenfei and BAI, LEI and Chen, Siheng}
}

@article{xu2026rethinking,
  title={Rethinking the Value of Multi-Agent Workflow: A Strong Single Agent Baseline},
  author={Xu, Jiawei and Koesdwiady, Arief and Bei, Sisong and Han, Yan and Huang, Baixiang and Wang, Dakuo and Chen, Yutong and Wang, Zheshen and Wang, Peihao and Li, Pan and others},
  journal={arXiv preprint arXiv:2601.12307},
  year={2026}
}

@article{dang2025multi,
  title={Multi-Agent Collaboration via Evolving Orchestration},
  author={Dang, Yufan and Qian, Chen and Luo, Xueheng and Fan, Jingru and Xie, Zihao and Shi, Ruijie and Chen, Weize and Yang, Cheng and Che, Xiaoyin and Tian, Ye and others},
  journal={arXiv preprint arXiv:2505.19591},
  year={2025}
}

@inproceedings{wang-etal-2025-megaagent,
    title = "{M}ega{A}gent: A Large-Scale Autonomous {LLM}-based Multi-Agent System Without Predefined {SOP}s",
    author = "Wang, Qian  and
      Wang, Tianyu  and
      Tang, Zhenheng  and
      Li, Qinbin  and
      Chen, Nuo  and
      Liang, Jingsheng  and
      He, Bingsheng",
    editor = "Che, Wanxiang  and
      Nabende, Joyce  and
      Shutova, Ekaterina  and
      Pilehvar, Mohammad Taher",
    booktitle = "Findings of the Association for Computational Linguistics: ACL 2025",
    month = jul,
    year = "2025",
    address = "Vienna, Austria",
    publisher = "Association for Computational Linguistics",
    url = "https://aclanthology.org/2025.findings-acl.259/",
    doi = "10.18653/v1/2025.findings-acl.259",
    pages = "4998--5036",
    ISBN = "979-8-89176-256-5"
}

@article{AlphaGeo,
  author = {Trieu H. Trinh and Yuhuai Wu and Quoc V. Le and He He and Thang Luong},
  title = {Solving olympiad geometry without human demonstrations},
  journal = {Nature},
  year = {2024},
  volume = {625},
  number = {7995},
  pages = {476--482},
  doi = {10.1038/s41586-023-06747-5},
  url = {https://doi.org/10.1038/s41586-023-06747-5},
}

@article{achiam2023gpt,
  title={Gpt-4 technical report},
  author={Achiam, Josh and Adler, Steven and Agarwal, Sandhini and Ahmad, Lama and Akkaya, Ilge and Aleman, Florencia Leoni and Almeida, Diogo and Altenschmidt, Janko and Altman, Sam and Anadkat, Shyamal and others},
  journal={arXiv preprint arXiv:2303.08774},
  year={2023}
}

@article{peng2023instruction,
  title={Instruction tuning with gpt-4},
  author={Peng, Baolin and Li, Chunyuan and He, Pengcheng and Galley, Michel and Gao, Jianfeng},
  journal={arXiv preprint arXiv:2304.03277},
  year={2023}
}

@article{chan2023chateval,
  title={Chateval: Towards better llm-based evaluators through multi-agent debate},
  author={Chan, Chi-Min and Chen, Weize and Su, Yusheng and Yu, Jianxuan and Xue, Wei and Zhang, Shanghang and Fu, Jie and Liu, Zhiyuan},
  journal={arXiv preprint arXiv:2308.07201},
  year={2023}
}

@inproceedings{chen2023agentverse,
  title={Agentverse: Facilitating multi-agent collaboration and exploring emergent behaviors},
  author={Chen, Weize and Su, Yusheng and Zuo, Jingwei and Yang, Cheng and Yuan, Chenfei and Chan, Chi-Min and Yu, Heyang and Lu, Yaxi and Hung, Yi-Hsin and Qian, Chen and others},
  booktitle={The Twelfth International Conference on Learning Representations},
  year={2023}
}

@inproceedings{nascimento2023self,
  title={Self-adaptive large language model (llm)-based multiagent systems},
  author={Nascimento, Nathalia and Alencar, Paulo and Cowan, Donald},
  booktitle={2023 IEEE International Conference on Autonomic Computing and Self-Organizing Systems Companion (ACSOS-C)},
  pages={104--109},
  year={2023},
  organization={IEEE}
}

@article{guo2025deepseek,
  title={Deepseek-r1: Incentivizing reasoning capability in llms via reinforcement learning},
  author={Guo, Daya and Yang, Dejian and Zhang, Haowei and Song, Junxiao and Zhang, Ruoyu and Xu, Runxin and Zhu, Qihao and Ma, Shirong and Wang, Peiyi and Bi, Xiao and others},
  journal={arXiv preprint arXiv:2501.12948},
  year={2025}
}

@article{wei2022chain,
  title={Chain-of-thought prompting elicits reasoning in large language models},
  author={Wei, Jason and Wang, Xuezhi and Schuurmans, Dale and Bosma, Maarten and Xia, Fei and Chi, Ed and Le, Quoc V and Zhou, Denny and others},
  journal={Advances in neural information processing systems},
  volume={35},
  pages={24824--24837},
  year={2022}
}

@article{singh2025agentic,
  title={Agentic reasoning and tool integration for llms via reinforcement learning},
  author={Singh, Joykirat and Magazine, Raghav and Pandya, Yash and Nambi, Akshay},
  journal={arXiv preprint arXiv:2505.01441},
  year={2025}
}

@article{shao2024deepseekmath,
  title={Deepseekmath: Pushing the limits of mathematical reasoning in open language models},
  author={Shao, Zhihong and Wang, Peiyi and Zhu, Qihao and Xu, Runxin and Song, Junxiao and Bi, Xiao and Zhang, Haowei and Zhang, Mingchuan and Li, YK and Wu, Yang and others},
  journal={arXiv preprint arXiv:2402.03300},
  year={2024}
}

@article{luo2025kbqa,
  title={Kbqa-o1: Agentic knowledge base question answering with monte carlo tree search},
  author={Luo, Haoran and Guo, Yikai and Lin, Qika and Wu, Xiaobao and Mu, Xinyu and Liu, Wenhao and Song, Meina and Zhu, Yifan and Tuan, Luu Anh and others},
  journal={arXiv preprint arXiv:2501.18922},
  year={2025}
}

@article{jin2025search,
  title={Search-r1: Training llms to reason and leverage search engines with reinforcement learning},
  author={Jin, Bowen and Zeng, Hansi and Yue, Zhenrui and Yoon, Jinsung and Arik, Sercan and Wang, Dong and Zamani, Hamed and Han, Jiawei},
  journal={arXiv preprint arXiv:2503.09516},
  year={2025}
}

@article{huan2025does,
  title={Does Math Reasoning Improve General LLM Capabilities? Understanding Transferability of LLM Reasoning},
  author={Huan, Maggie and Li, Yuetai and Zheng, Tuney and Xu, Xiaoyu and Kim, Seungone and Du, Minxin and Poovendran, Radha and Neubig, Graham and Yue, Xiang},
  journal={arXiv preprint arXiv:2507.00432},
  year={2025}
}

@article{xiao2025trading,
  title={Trading-r1: Financial trading with llm reasoning via reinforcement learning},
  author={Xiao, Yijia and Sun, Edward and Chen, Tong and Wu, Fang and Luo, Di and Wang, Wei},
  journal={arXiv preprint arXiv:2509.11420},
  year={2025}
}

@article{zhang2025mars,
  title={Mars: A multi-agent framework incorporating socratic guidance for automated prompt optimization},
  author={Zhang, Jian and Wang, Zhangqi and Zhu, Haiping and Liu, Jun and Lin, Qika and Cambria, Erik},
  journal={arXiv preprint arXiv:2503.16874},
  year={2025}
}

@inproceedings{cai-etal-2025-unilaw,
    title = "Unilaw-R1: A Large Language Model for Legal Reasoning with Reinforcement Learning and Iterative Inference",
    author = "Cai, Hua  and
      Zhao, Shuang  and
      Zhang, Liang  and
      Shen, Xuli  and
      Xu, Qing  and
      Shen, Weilin  and
      Wen, Zihao  and
      Ban, Tianke",
    editor = "Christodoulopoulos, Christos  and
      Chakraborty, Tanmoy  and
      Rose, Carolyn  and
      Peng, Violet",
    booktitle = "Proceedings of the 2025 Conference on Empirical Methods in Natural Language Processing",
    month = nov,
    year = "2025",
    address = "Suzhou, China",
    publisher = "Association for Computational Linguistics",
    url = "https://aclanthology.org/2025.emnlp-main.915/",
    doi = "10.18653/v1/2025.emnlp-main.915",
    pages = "18117--18131",
    ISBN = "979-8-89176-332-6"
}

@article{yan2025memory,
  title={Memory-r1: Enhancing large language model agents to manage and utilize memories via reinforcement learning},
  author={Yan, Sikuan and Yang, Xiufeng and Huang, Zuchao and Nie, Ercong and Ding, Zifeng and Li, Zonggen and Ma, Xiaowen and Kersting, Kristian and Pan, Jeff Z and Sch{\"u}tze, Hinrich and others},
  journal={arXiv preprint arXiv:2508.19828},
  year={2025}
}

@article{huang2025mobilevla,
  title={Mobilevla-r1: Reinforcing vision-language-action for mobile robots},
  author={Huang, Ting and Li, Dongjian and Yang, Rui and Zhang, Zeyu and Yang, Zida and Tang, Hao},
  journal={arXiv preprint arXiv:2511.17889},
  year={2025}
}

@article{wei2025webagent,
  title={Webagent-r1: Training web agents via end-to-end multi-turn reinforcement learning},
  author={Wei, Zhepei and Yao, Wenlin and Liu, Yao and Zhang, Weizhi and Lu, Qin and Qiu, Liang and Yu, Changlong and Xu, Puyang and Zhang, Chao and Yin, Bing and others},
  journal={arXiv preprint arXiv:2505.16421},
  year={2025}
}

@article{luo2025hypergraphrag,
  title={HyperGraphRAG: Retrieval-Augmented Generation via Hypergraph-Structured Knowledge Representation},
  author={Luo, Haoran and Chen, Guanting and Zheng, Yandan and Wu, Xiaobao and Guo, Yikai and Lin, Qika and Feng, Yu and Kuang, Zemin and Song, Meina and Zhu, Yifan and others},
  journal={arXiv preprint arXiv:2503.21322},
  year={2025}
}

@inproceedings{ho2020constructing,
  title        = {Constructing A Multi-hop QA Dataset for Comprehensive Evaluation of Reasoning Steps},
  author       = {Xanh Ho and Anh-Khoa Duong Nguyen and Saku Sugawara and Akiko Aizawa},
  booktitle    = {Proceedings of the 28th International Conference on Computational Linguistics (COLING 2020)},
  pages        = {6609--6625},
  year         = {2020},
  publisher    = {International Committee on Computational Linguistics},
  url          = {https://aclanthology.org/2020.coling-main.580},
  doi          = {10.18653/v1/2020.coling-main.580}
}

@inproceedings{yang-etal-2018-hotpotqa,
    title = "{H}otpot{QA}: A Dataset for Diverse, Explainable Multi-hop Question Answering",
    author = "Yang, Zhilin  and
      Qi, Peng  and
      Zhang, Saizheng  and
      Bengio, Yoshua  and
      Cohen, William  and
      Salakhutdinov, Ruslan  and
      Manning, Christopher D.",
    editor = "Riloff, Ellen  and
      Chiang, David  and
      Hockenmaier, Julia  and
      Tsujii, Jun{'}ichi",
    booktitle = "Proceedings of the 2018 Conference on Empirical Methods in Natural Language Processing",
    month = oct # "-" # nov,
    year = "2018",
    address = "Brussels, Belgium",
    publisher = "Association for Computational Linguistics",
    url = "https://aclanthology.org/D18-1259/",
    doi = "10.18653/v1/D18-1259",
    pages = "2369--2380"
}

@article{cobbe2021training,
  title={Training verifiers to solve math word problems},
  author={Cobbe, Karl and Kosaraju, Vineet and Bavarian, Mohammad and Chen, Mark and Jun, Heewoo and Kaiser, Lukasz and Plappert, Matthias and Tworek, Jerry and Hilton, Jacob and Nakano, Reiichiro and others},
  journal={arXiv preprint arXiv:2110.14168},
  year={2021}
}

@article{yu2025dapo,
  title={Dapo: An open-source llm reinforcement learning system at scale},
  author={Yu, Qiying and Zhang, Zheng and Zhu, Ruofei and Yuan, Yufeng and Zuo, Xiaochen and Yue, Yu and Dai, Weinan and Fan, Tiantian and Liu, Gaohong and Liu, Lingjun and others},
  journal={arXiv preprint arXiv:2503.14476},
  year={2025}
}

@article{trivedi2022musique,
  title={MuSiQue: Multihop Questions via Single-hop Question Composition},
  author={Trivedi, Harsh and Balasubramanian, Niranjan and Khot, Tushar and Sabharwal, Ashish},
  journal={Transactions of the Association for Computational Linguistics},
  volume={10},
  pages={539--554},
  year={2022},
  publisher={MIT Press One Broadway, 12th Floor, Cambridge, Massachusetts 02142, USA~…}
}

@inproceedings{mallen-etal-2023-trust,
    title = "When Not to Trust Language Models: Investigating Effectiveness of Parametric and Non-Parametric Memories",
    author = "Mallen, Alex  and
      Asai, Akari  and
      Zhong, Victor  and
      Das, Rajarshi  and
      Khashabi, Daniel  and
      Hajishirzi, Hannaneh",
    editor = "Rogers, Anna  and
      Boyd-Graber, Jordan  and
      Okazaki, Naoaki",
    booktitle = "Proceedings of the 61st Annual Meeting of the Association for Computational Linguistics (Volume 1: Long Papers)",
    month = jul,
    year = "2023",
    address = "Toronto, Canada",
    publisher = "Association for Computational Linguistics",
    url = "https://aclanthology.org/2023.acl-long.546/",
    doi = "10.18653/v1/2023.acl-long.546",
    pages = "9802--9822"
}

@inproceedings{kryscinski-etal-2022-booksum,
    title = "{BOOKSUM}: A Collection of Datasets for Long-form Narrative Summarization",
    author = "Kryscinski, Wojciech  and
      Rajani, Nazneen  and
      Agarwal, Divyansh  and
      Xiong, Caiming  and
      Radev, Dragomir",
    editor = "Goldberg, Yoav  and
      Kozareva, Zornitsa  and
      Zhang, Yue",
    booktitle = "Findings of the Association for Computational Linguistics: EMNLP 2022",
    month = dec,
    year = "2022",
    address = "Abu Dhabi, United Arab Emirates",
    publisher = "Association for Computational Linguistics",
    url = "https://aclanthology.org/2022.findings-emnlp.488/",
    doi = "10.18653/v1/2022.findings-emnlp.488",
    pages = "6536--6558"
}

@article{huang2024gpt,
  title={The gpt-writingprompts dataset: A comparative analysis of character portrayal in short stories},
  author={Huang, Xi Yu and Vishnubhotla, Krishnapriya and Rudzicz, Frank},
  journal={arXiv preprint arXiv:2406.16767},
  year={2024}
}

@inproceedings{amini-etal-2019-mathqa,
    title = "{M}ath{QA}: Towards Interpretable Math Word Problem Solving with Operation-Based Formalisms",
    author = "Amini, Aida  and
      Gabriel, Saadia  and
      Lin, Shanchuan  and
      Koncel-Kedziorski, Rik  and
      Choi, Yejin  and
      Hajishirzi, Hannaneh",
    editor = "Burstein, Jill  and
      Doran, Christy  and
      Solorio, Thamar",
    booktitle = "Proceedings of the 2019 Conference of the North {A}merican Chapter of the Association for Computational Linguistics: Human Language Technologies, Volume 1 (Long and Short Papers)",
    month = jun,
    year = "2019",
    address = "Minneapolis, Minnesota",
    publisher = "Association for Computational Linguistics",
    url = "https://aclanthology.org/N19-1245/",
    doi = "10.18653/v1/N19-1245",
    pages = "2357--2367"
}

@article{rajpurkar2018know,
  title={Know what you don't know: Unanswerable questions for SQuAD},
  author={Rajpurkar, Pranav and Jia, Robin and Liang, Percy},
  journal={arXiv preprint arXiv:1806.03822},
  year={2018}
}

@article{joshi2017triviaqa,
  title={Triviaqa: A large scale distantly supervised challenge dataset for reading comprehension},
  author={Joshi, Mandar and Choi, Eunsol and Weld, Daniel S and Zettlemoyer, Luke},
  journal={arXiv preprint arXiv:1705.03551},
  year={2017}
}

@article{agrawal2025gepa,
  title={Gepa: Reflective prompt evolution can outperform reinforcement learning},
  author={Agrawal, Lakshya A and Tan, Shangyin and Soylu, Dilara and Ziems, Noah and Khare, Rishi and Opsahl-Ong, Krista and Singhvi, Arnav and Shandilya, Herumb and Ryan, Michael J and Jiang, Meng and others},
  journal={arXiv preprint arXiv:2507.19457},
  year={2025}
}

@article{yuksekgonul2024textgrad,
  title={Textgrad: Automatic" differentiation" via text},
  author={Yuksekgonul, Mert and Bianchi, Federico and Boen, Joseph and Liu, Sheng and Huang, Zhi and Guestrin, Carlos and Zou, James},
  journal={arXiv preprint arXiv:2406.07496},
  year={2024}
}

@inproceedings{yang2024llm_optimizers,
  title        = {Large Language Models as Optimizers},
  author       = {Chengrun Yang and Xuezhi Wang and Yifeng Lu and Hanxiao Liu and Quoc V. Le and Denny Zhou and Xinyun Chen},
  booktitle    = {Proceedings of the International Conference on Learning Representations (ICLR) 2024},
  year         = {2024},
  url          = {https://openreview.net/forum?id=Bb4VGOWELI},
  note         = {OpenReview preprint},
}

\newpage
\appendix
\section{Sample Prompt}
\label{sec:sample-prompt}

This section provides concrete prompting examples used in OrchMAS and explains, in a
case-study style, why the multi-agent collaboration produces \emph{more reliable}
multi-hop reasoning than a single-pass LLM response. We use the 2WikiMultihopQA
cases in Table~\ref{tab:case-study-3} and Table~\ref{tab:case-study-combined} as
running examples. In particular, these cases highlight a common failure mode of
vanilla LLM prompting: \textbf{premature commitment} (answering before verifying
latent entities) and \textbf{entity alias confusion} (incorrectly mapping a query
to a more salient but irrelevant entity). OrchMAS mitigates both by decomposing the
task into role-specialized turns, and by explicitly separating \emph{evidence
acquisition} from \emph{final decision}.

\noindent\textbf{Overall prompting principle.}
Given an input question $x$, the coordinator $\mathcal{C}$ constructs a
collaboration trajectory by dynamically assigning roles $\sigma_t$ and producing
messages $u_t^{\mathrm{msg}}$ along with concise rationales
$u_t^{\mathrm{reason}}$. Each agent returns feedback $v_t^{\mathrm{msg}}$ that is
treated as \emph{intermediate evidence} rather than the final answer. The final
role (\texttt{an assistant}) must only output the answer after the evidence is
sufficiently consistent across turns.

\subsection{Coordinator-to-Agent Prompt Template}
\label{subsec:prompt-template}

We show a representative prompt template for each role. In practice, OrchMAS uses
structured role instructions and minimal cross-role leakage to prevent an agent
from bypassing its responsibility (e.g., forcing the \texttt{researcher} to
retrieve and justify, and forcing the \texttt{clarifier} to verify ambiguities).

\noindent\textbf{(1) Researcher prompt (evidence acquisition).}
\begin{quote}
\small
\textbf{System:} You are a \texttt{researcher} agent in a multi-agent reasoning
system. Your job is to identify missing facts needed to answer the question.
Do \emph{not} finalize the answer. Instead, return (i) the key entities, (ii) the
facts to retrieve/verify, and (iii) a short evidence-based conclusion.

\textbf{User:} Question: \{INPUT\_QUESTION\}. \\
You must: \\
1) Extract the latent sub-questions required to answer. \\
2) Provide the minimal factual chain needed (entity $\rightarrow$ attribute
$\rightarrow$ comparison). \\
3) Return your findings in the format: \\
\texttt{Entities: ...} \\
\texttt{Facts: ...} \\
\texttt{Conclusion (not final answer): ...}
\end{quote}

\noindent\textbf{(2) Clarifier prompt (disambiguation and contradiction check).}
\begin{quote}
\small
\textbf{System:} You are a \texttt{clarifier} agent. Your job is to detect
ambiguities, name collisions, or contradictions in the \texttt{researcher}
output. Do \emph{not} finalize the answer. If there is any plausible ambiguity
(e.g., multiple artists sharing a song title), you must explicitly verify which
entity the question refers to.

\textbf{User:} Question: \{INPUT\_QUESTION\}. \\
Researcher evidence: \{RESEARCHER\_OUTPUT\}. \\
You must: \\
1) Identify potential ambiguity points (title collisions, multiple candidates,
country vs birthplace confusion, etc.). \\
2) Propose verification steps and resolve the ambiguity. \\
3) Return: \\
\texttt{Ambiguities: ...} \\
\texttt{Verification: ...} \\
\texttt{Resolved evidence: ...}
\end{quote}

\noindent\textbf{(3) Assistant prompt (decision and concise final output).}
\begin{quote}
\small
\textbf{System:} You are \texttt{an assistant}. Output only the final answer.
You must not add extra explanation beyond what is asked. You must base the final
answer strictly on the provided evidence. If evidence conflicts, ask for one
more verification step (but in our evaluation setting, you must still return the
best-supported answer).

\textbf{User:} Question: \{INPUT\_QUESTION\}. \\
Evidence from prior turns: \{EVIDENCE\_PACKET\}. \\
Return: \texttt{Answer: <short answer>}
\end{quote}

\subsection{Why the Reasoning is Rational and Necessary}
\label{subsec:why-reasoning-rational}

\paragraph{Case: nationality comparison (Table~\ref{tab:case-study-3}).}
The question asks whether two films have directors from the \emph{same country},
which implicitly requires: (i) identifying each film's director, and (ii)
mapping each director to a country-level attribute (nationality or country of
origin). A single LLM, when prompted in one shot, often fails here due to two
typical behaviors:

\begin{itemize}
\item \textbf{Premature commitment:} the model may guess a director based on
surface familiarity (e.g., confusing similarly titled films or assuming a
well-known director), then propagate the guessed nationality into the final
comparison without verification.
\item \textbf{Attribute mismatch:} the model may conflate \emph{birthplace},
\emph{residence}, and \emph{nationality} when the question strictly requires a
country identity. This is especially common in multi-hop settings where the
model compresses multiple steps into a single narrative and accidentally swaps
the attribute used for comparison.
\end{itemize}

OrchMAS makes the reasoning explicit and staged: the \texttt{researcher} first
retrieves the director identities and country signals, producing a localized and
auditable claim (\textit{Director A $\rightarrow$ Country A; Director B
$\rightarrow$ Country B}) as shown in Turn~1 of
Table~\ref{tab:case-study-3}. Only after this evidence is present does
\texttt{an assistant} emit the final decision (Turn~2). This separation is
rational because the final answer is merely a deterministic comparison once the
two country attributes are established. In other words, OrchMAS forces the model to
treat the problem as: \emph{retrieve $\rightarrow$ normalize attribute
$\rightarrow$ compare}, rather than \emph{guess $\rightarrow$ justify}.

\paragraph{Case: song-performer nationality (Table~\ref{tab:case-study-combined}, Case 2).}
This case demonstrates a more subtle and more realistic failure mode: \textbf{name
collision and salience bias}. The query ``You're My One And Only Love'' is a
title that may be associated with multiple performers across genres and eras.
A single LLM is prone to anchor on the most salient candidate it recalls (or the
most frequent co-occurrence in its training distribution), then outputs a
nationality consistent with that guess. This is exactly the scenario where
a one-pass answer is likely to be wrong: the model's top-of-mind association
(\emph{e.g., a famous singer}) can differ from the dataset's intended entity.

OrchMAS addresses this by making \emph{disambiguation} a first-class objective of
the \texttt{clarifier}. In Table~\ref{tab:case-study-combined} (Case 2),
Turn~1 provides a tentative nationality claim. However, the \texttt{clarifier}
explicitly flags a contradiction with an alternative plausible performer and
forces a verification step (Turn~2). This yields an evidence-grounded resolution
of the actual artist identity, after which the final answer becomes stable.
Therefore, the reasoning is not only ``reasonable'' but \emph{necessary}:
without disambiguation, the problem is under-specified from the model's internal
memory alone.

\subsection{How MAS Turns a Likely Wrong Single-LLM Answer into a Correct One}
\label{subsec:why-mas-works}

We now summarize the mechanism-level explanation for why OrchMAS can correct errors
that a single LLM tends to make.

\paragraph{(i) Role separation reduces shortcut learning in inference.}
In single-LLM prompting, the model is rewarded (implicitly) for producing a
fluent final answer quickly, which encourages shortcut heuristics such as
choosing the most salient entity and generating a plausible justification. OrchMAS
removes this incentive by \emph{denying the researcher the ability to finalize}.
The \texttt{researcher} must output structured intermediate facts, which shifts
the model behavior from ``answering'' to ``retrieving + stating.'' This reduces
hallucination probability because intermediate outputs are constrained to
verifiable attributes (director name, birthplace, nationality).

\paragraph{(ii) Explicit contradiction checking prevents early-stage error propagation.}
A single wrong entity choice in Turn~1 of a one-pass solution contaminates all
downstream reasoning. In OrchMAS, the \texttt{clarifier} is specialized to catch
exactly this: ambiguous entity mentions, conflicts between candidates, and
attribute mismatch. This is visible in Table~\ref{tab:case-study-combined}
(Case 2), where the clarifier refuses to accept the initial guess and requires
verification of the artist identity. Practically, this means OrchMAS adds a
``circuit breaker'' stage: it is cheaper to correct the entity before composing
the final answer than to patch an already-committed final narrative.

\paragraph{(iii) Evidence-first decision makes the final step trivial and robust.}
Once evidence is assembled, the final step often reduces to a simple operation:
\emph{string match}, \emph{location extraction}, or \emph{equality/inequality
comparison}. This is the key reason OrchMAS improves reliability: it pushes the
uncertainty into earlier turns (where verification is allowed) and makes the
final stage a low-variance mapping. For example, in
Table~\ref{tab:case-study-3}, once (Taiwan) and (Germany) are obtained, the
answer ``No'' follows deterministically.

\paragraph{(iv) Multi-turn structure aligns with the latent multi-hop structure of the dataset.}
2WikiMultihopQA questions are constructed to require multiple supporting facts.
A one-pass LLM response must internally simulate multi-hop retrieval, but it may
collapse steps or omit a hop, especially when two different entities must be
linked and compared. OrchMAS externalizes these hops into turns, aligning the
reasoning trace with the dataset's compositional structure (entity identification
$\rightarrow$ attribute retrieval $\rightarrow$ comparison). This alignment
reduces the chance that the model implicitly ``fills gaps'' with plausible but
incorrect information.
\begin{table*}[!t]
\centering
\caption{Case study on 2WikiMultihopQA (Data ID: 3). The coordinator $\mathcal{C}$ dynamically assigns agent roles $\sigma_t$ to resolve a nationality comparison question. Ground truth: \textit{No}.}
\label{tab:case-study-3}
\small
\renewcommand{\arraystretch}{1.3}
\begin{tabular}{p{1.2cm} p{1.5cm} p{4.5cm} p{4.0cm} p{4.0cm}}
\toprule
\textbf{Turn} & \textbf{Role $\sigma_t$} & \textbf{Reasoning $u_t^{\mathrm{reason}}$} & \textbf{Request $u_t^{\mathrm{msg}}$} & \textbf{Feedback $v_t^{\mathrm{msg}}$} \\
\midrule
\multicolumn{5}{l}{\textbf{Question:} \textit{Do both Beauty and Bullets and In the Name of the King 3: The Last Mission films have the}} \\
\multicolumn{5}{l}{\textit{directors from the same country?}} \\
\midrule
Turn 1 & \texttt{researcher} &
The question asks whether the directors of ``Beauty and Bullets'' and ``In the Name of the King 3: The Last Mission'' are from the same country. I need to identify each film's director and their nationality. &
Do both Beauty and Bullets and In the Name of the King 3: The Last Mission films have the directors from the same country? &
\textbf{Beauty and Bullets}: directed by Derrick Ku (China). \textbf{In the Name of the King 3}: directed by Uwe Boll (Germany). China $\neq$ Germany, so the directors do \textbf{not} share the same country. \\
\midrule
Turn 2 & \texttt{an assistant} &
The response confirms that the two directors are from different countries (China vs.\ Germany). The answer is definitively ``No''. No further verification needed. &
\multicolumn{2}{l}{\textbf{Answer:} No} \\
\bottomrule
\end{tabular}
\label{Two_turns}
\end{table*}

\begin{table*}[!t]
\centering
\caption{Case studies on 2WikiMultihopQA. The coordinator $\mathcal{C}$ dynamically assigns agent roles $\sigma_t$ at each turn. Both cases demonstrate the three-phase collaboration: \texttt{researcher} $\rightarrow$ \texttt{clarifier} $\rightarrow$ \texttt{an assistant}.}
\label{tab:case-study-combined}
\small
\renewcommand{\arraystretch}{1.3}
\begin{tabular}{p{0.8cm} p{1.4cm} p{4.2cm} p{3.8cm} p{4.0cm}}
\toprule
\textbf{Turn} & \textbf{Role $\sigma_t$} & \textbf{Reasoning $u_t^{\mathrm{reason}}$} & \textbf{Request $u_t^{\mathrm{msg}}$} & \textbf{Feedback $v_t^{\mathrm{msg}}$} \\

\midrule
\multicolumn{5}{l}{\textbf{Case 1:} \textit{Where was the director of film A Zed \& Two Noughts born?} \hfill Ground truth: \textit{Newport, Wales}} \\
\midrule
1 & \texttt{researcher} &
The question asks about the birthplace of the director of ``A Zed \& Two Noughts.'' I need to identify the director first. &
Who was the director of the film ``A Zed \& Two Noughts''? &
The film was directed by \textbf{Peter Greenaway}. \\
\cmidrule(l){1-5}
2 & \texttt{clarifier} &
The director is Peter Greenaway. Now I need to find out where he was born. &
Where was Peter Greenaway born? &
Peter Greenaway was born in \textbf{Newport, Monmouthshire, Wales, UK} (5 April 1942). \\
\cmidrule(l){1-5}
3 & \texttt{an assistant} &
Peter Greenaway was born in Newport, Monmouthshire, Wales. The answer is confirmed. &
\multicolumn{2}{l}{\textbf{Answer:} Newport, Monmouthshire, Wales \checkmark} \\

\midrule
\multicolumn{5}{l}{\textbf{Case 2:} \textit{What nationality is the performer of song You're My One And Only Love?} \hfill Ground truth: \textit{American}} \\
\midrule
1 & \texttt{researcher} &
The question asks the nationality of the performer. I suspect Celine Dion (Canadian), but need to verify the actual performer. &
What nationality is the performer of song ``You're My One And Only Love''? &
The performer is \textbf{American}. \\
\cmidrule(l){1-5}
2 & \texttt{clarifier} &
The response says American, contradicting my initial guess of Celine Dion. I need to verify the exact artist. &
Verify the nationality of the artist for the song ``You're My One And Only Love.'' &
The primary artist is \textbf{Jimmy Brown} (jazz vocalist), born in Chicago, Illinois, USA. Nationality: \textbf{American}. \\
\cmidrule(l){1-5}
3 & \texttt{an assistant} &
Confirmed: the song is by Jimmy Brown (American), not Celine Dion. The correct answer is American. &
\multicolumn{2}{l}{\textbf{Answer:} American \checkmark} \\
\bottomrule
\end{tabular}
\label{Three_turns}
\end{table*}
\subsection{Prompt Instantiation for the Case Study Tables}
\label{subsec:prompt-instantiation}

We finally show how the prompts instantiate on the case study in
Table~\ref{tab:case-study-3}. The coordinator $\mathcal{C}$ selects
\texttt{researcher} at Turn~1 because the question requires evidence gathering
(two directors + two countries). After receiving the evidence, $\mathcal{C}$
skips \texttt{clarifier} and directly assigns \texttt{an assistant} at Turn~2,
because the evidence is already non-ambiguous (two distinct countries). This
demonstrates \textbf{dynamic depth control}: OrchMAS does not force a fixed
three-stage pipeline; instead, it conditionally inserts clarification only when
ambiguity is detected.

Similarly, in Table~\ref{tab:case-study-combined}, OrchMAS uses the full
three-phase pattern in both cases, but for different reasons: Case~1 uses
\texttt{clarifier} to perform a clean attribute hop (director $\rightarrow$
birthplace), while Case~2 uses \texttt{clarifier} to resolve an entity collision
(performer identity). These examples illustrate that OrchMAS's prompts are designed
around \emph{error modes}: when the risk is ``missing hop,'' clarification is a
structured second hop; when the risk is ``wrong entity,'' clarification becomes
a verification gate.

\paragraph{Takeaway.}
The correctness gains from MAS are not merely due to ``more tokens'' or ``more
steps,'' but due to a disciplined division of labor: \textbf{retrieval and
disambiguation are made explicit responsibilities}, and the final answer is
withheld until the evidence is consistent. This is precisely why OrchMAS can turn a
likely incorrect one-shot response into a correct multi-hop decision in
case-study settings like Table~\ref{tab:case-study-3} and
Table~\ref{tab:case-study-combined}.

\end{document}